%% file: main.tex
\documentclass{article}

\usepackage[main,preprint]{neurips_2026}

\usepackage[utf8]{inputenc}
\usepackage[T1]{fontenc}
\usepackage[colorlinks=true,linkcolor=blue,citecolor=blue,urlcolor=blue]{hyperref}
\usepackage{url}
\usepackage{booktabs}
\usepackage{amsmath}
\usepackage{amssymb}
\usepackage{graphicx}
\usepackage{enumitem}
\usepackage{multirow}
\usepackage{xcolor}
\usepackage{comment}
\input{results_macros}

\title{PHREEQC-MCQ-200: A Diagnostic Benchmark for Tool-Augmented Scientific Simulator Agents}

\author{
Ke Zhang\\
University of California, Riverside\\
\texttt{kzhan153@ucr.edu}
\And
Sahchit Chundur\\
University of California, Irvine\\
\texttt{smchundu@uci.edu}
\And
Mohammad Javad Qomi\thanks{Corresponding authors.}\\
University of California, Irvine\\
\texttt{mjaq@uci.edu}
\And
Maziar Raissi\footnotemark[1]\\
University of California, Riverside\\
\texttt{maziar.raissi1@ucr.edu}
}

\begin{document}

\maketitle
\begin{abstract}
Large language model agents are increasingly connected to scientific software, yet it remains unclear when tool access makes scientific computation more reliable rather than merely more complex. We introduce PHREEQC-MCQ-200, a benchmark for evaluating tool-augmented agents on deterministic aqueous-geochemistry simulations. The benchmark contains 200 multiple-choice questions derived from 21 validated PHREEQC scenarios, requiring agents to construct simulator inputs, execute PHREEQC, inspect structured outputs, and commit to final answers.

Across multiple frontier and mid-tier model families, simulator access substantially improves aggregate accuracy, confirming that grounded execution is necessary for many scientific-computation tasks. However, the gains are not monotonic: tool-augmented agents also lose items they answered correctly without tools, revealing regressions that average accuracy alone hides. We further show that output-access protocol matters. A table-of-contents interface can reduce token cost while preserving or improving accuracy for stronger models, but it degrades performance for mid-tier models that cannot reliably navigate structured simulator outputs.

PHREEQC-MCQ-200 therefore frames scientific tool use as an end-to-end diagnostic problem rather than a simple tool-calling capability. We argue that evaluations of scientific agents should report not only accuracy, but also item-level retention, output-access sensitivity, trajectory failures, and where the computation chain breaks.
\end{abstract}

\section{Introduction}

Building tool-augmented LLM agents has become routine; evaluating whether tools actually make them more reliable on real scientific work has not. LLM agents are increasingly connected to external tools, including calculators, databases, code interpreters, and scientific software. Frameworks such as ReAct \citep{yao2023react} and Toolformer \citep{schick2023toolformer} helped establish tool use as a core language-model capability, and scientific systems such as ChemCrow \citep{bran2024chemcrow} and autonomous chemistry agents \citep{boiko2023autonomous} show that tools can extend model behavior beyond text-only prediction. For scientific computation, however, the central evaluation question is not simply whether a model can call a tool. It is when tool augmentation improves a model's ability to operate a deterministic scientific workflow, and when the tool loop itself becomes a source of errors~\citep{toolusetax2026}. We study this on a deterministic scientific simulator and find a cross-vendor capability-tier $\times$ tool-use interaction.

Deterministic scientific simulators make this question testable: simulator tasks provide executable evidence and reproducible ground truth. We study this through PHREEQC, an open-source simulator for aqueous geochemistry widely used by researchers and environmental consultants \citep{parkhurst2013phreeqc}. PHREEQC supports equilibrium speciation, mineral saturation, surface complexation, kinetic reactions, and one-dimensional reactive transport \citep{parkhurst2013phreeqc}. Its application scope spans groundwater and pollution studies \citep{appelo2004geochemistry}, geothermal-well scaling and corrosion \citep{bozau2015hydrogeochemical}, and reactive-transport workflows \citep{parkhurst2013phreeqc}. The geothermal setting makes the practical stakes concrete: market estimates project global geothermal revenue approaching \$10 billion by 2030 \citep{grandview2024geothermalmarket}, while PHREEQC has been used to model brine chemistry, mineral scaling, and corrosion risks that affect whether subsurface-energy workflows remain operable \citep{bozau2015hydrogeochemical}. PHREEQC is deterministic for a fixed input and thermodynamic database, a precondition for a reproducible benchmark that the leading commercial geochemical-modeling alternatives do not match. Its section-structured text outputs, shared by many command-line scientific tools (electronic-structure, molecular-simulation, subsurface-flow), make it a realistic case for studying scientific-tool use without requiring reviewers to evaluate novelty in geochemistry.

We introduce \BenchName, a high-confidence benchmark of \BenchN{} multiple-choice question items derived from \BenchScenarioN{} validated simulation scenarios. Each item is answerable by constructing and executing a PHREEQC simulation and extracting a specific computed quantity. The benchmark uses multiple choice because it provides deterministic scalable grading, but every item is grounded in a unique simulator-derived target value. We therefore use the benchmark as a diagnostic harness for evaluating scientific agents, not as a geochemistry trivia set.

The empirical surprise is that simulator access is not a monotone intervention: it gains and loses items simultaneously (Section~\ref{sec:diag-retention}). We measure this with a kept/gained/lost decomposition --- connecting item-level gain/loss accounting from model-version-comparison work~\citep{rci2026} with tool-vs-no-tool evaluation --- and use it alongside chain-of-thought controls, output-access ablations, trajectory metrics, and failure analysis as the central diagnostics of the benchmark.

This paper asks five evaluation questions:
\begin{enumerate}[leftmargin=*, itemsep=2pt]
    \item Can scaling chain-of-thought reasoning substitute for grounded simulator execution, or is tool access irreducibly necessary?
    \item When tool use lifts aggregate accuracy, what items does the agent quietly lose in exchange?
    \item Can a metadata-guided interface let agents navigate large simulator outputs at a fraction of the input-token cost without losing accuracy?
    \item Do output-access tradeoffs and tool-induced regressions track model capability tier, or vendor identity?
    \item Where does the externally observable computation chain break: input construction, execution, output navigation, or final-answer mapping?
\end{enumerate}

Our contribution is an evaluation benchmark and analysis methodology, not a new geochemical model or a new agent architecture. We provide one benchmark and five findings, paired one-to-one with the five evaluation questions above:
\begin{enumerate}[leftmargin=*, itemsep=2pt]
    \item We provide \BenchName, an expert-validated benchmark and artifact package for evaluating LLM agents in a deterministic, unique-answer scientific software setting.
    \item \textbf{(Q1)} Scaling reasoning tokens cannot substitute for tool access: chain-of-thought yields $\pm 8$\,pp at 29--140$\times$ output-token expansion, while simulator access adds 15.0--41.5\,pp on top of the better no-tool baseline for top- and mid-tier agents.
    \item \textbf{(Q2)} A gain/loss/retention decomposition showing that tool augmentation is never purely additive: top- and mid-tier agents lose 10--32 items they had answered correctly without tools while gaining 62--110 newly correct ones, yielding retention rates of 56.4--86.5\%. This connects item-level gain/loss accounting from model-version-comparison work~\citep{rci2026} with tool-vs-no-tool evaluation.
    \item \textbf{(Q3)} A within-family output-access protocol comparison (table-of-contents indexing vs truncated raw output), showing 11--57\% input-token reduction while matching or exceeding raw-output accuracy by 0.5--2.5\,pp for top-tier models, with the comparison validated by 3 independent TOC reruns per model on all five top- and mid-tier agents (top-tier rerun SDs 0.6--2.8\,pp; max 4.1\,pp on GPT-5.1). This instantiates tool-output-format ablation~\citep{toolouts2025} for scientific simulator outputs.
    \item \textbf{(Q4)} Our central empirical finding: output-access tradeoffs cluster by model capability tier and cross vendor boundaries --- on the output-access axis, GPT-5.4 patterns with Claude Opus 4.6 and Sonnet 4.6 (TOC matches or exceeds Raw100k by 0.5--2.5\,pp at 11--57\% input-token reduction), while same-vendor GPT-5.1 and GPT-5.2 lose 7.5--9.5\,pp under TOC. Retention sits intermediate for GPT-5.4 (64.8\%, between Sonnet's 86.5\% and the mid-tier GPT 56--59\% range), so the cross-vendor effect is axis-specific --- concentrated where output-access scaffolding interacts with capability. To our knowledge this is the first such finding in a controlled scientific-simulator setting.
    \item \textbf{(Q5)} We report a domain-expert failure taxonomy and an external-harness comparison (Claude Code CLI, Codex CLI, matched 50-question subsample) showing 2.7--3.3$\times$ better tokens-per-correct-answer for our custom harness (Appendix~\ref{sec:appendix-external-harness}).
\end{enumerate}

\section{Related Work}

\paragraph{Tool-augmented language models.}
Prior work has studied agents that interleave language-model reasoning with actions or external calls \citep{yao2023react,schick2023toolformer}. Scientific tool-use systems demonstrate the promise of connecting LLMs to chemistry tools and laboratory or simulation workflows \citep{bran2024chemcrow,boiko2023autonomous}. Our focus is narrower and more diagnostic: rather than proposing a general autonomous scientist, we evaluate when simulator access improves or degrades performance under controlled grading.

\paragraph{Benchmarks for agentic systems.}
LLM evaluation has moved from static QA benchmarks toward agent benchmarks that grade executable artifacts and tool use \citep{rein2023gpqa,jimenez2024swebench}. In AI for science the same shift is underway: ChemBench probes chemistry knowledge against expert chemists \citep{mirza2025chembench}, while ScienceAgentBench evaluates generated programs and execution results for data-driven discovery \citep{chen2025scienceagentbench}. \BenchName{} fills a gap between these: it grades agents that operate a real \emph{deterministic scientific simulator} end-to-end --- input synthesis, execution, output navigation, and final-answer commitment --- with simulator-derived ground truth rather than program-output equivalence or expert preference. Its distinctive contribution is a PHREEQC-specific benchmark and artifact package with simulator inputs, outputs, trajectories, output-access ablations, and failure-analysis queues.

\paragraph{Tool-use cost and output-access ablations.}
A line of recent work decomposes the cost of tool-augmented reasoning into protocol- and execution-level components and shows that tool calls are not free of error \citep{toolusetax2026}, evaluates how LLMs process structured tool outputs across return formats \citep{toolouts2025}, and applies item-level reliable-change detection to surface bidirectional movement masked by aggregate accuracy \citep{rci2026}. \BenchName{} integrates these moves on a deterministic scientific simulator: we ablate output-access protocol (TOC vs.\ truncated raw output), apply gain/loss/retention accounting to tool vs.\ no-tool comparisons, and report per-model cost alongside accuracy.

\section{Benchmark and Agent Harness}
\label{sec:benchmark}

\subsection{Dataset and Task}

\BenchName{} contains \BenchN{} four-option multiple-choice items derived from \BenchScenarioN{} validated PHREEQC simulation scenarios authored and reviewed by domain experts in aqueous geochemistry, with topic ideas inspired by the official PHREEQC examples and the phreeqcusers community forum. Each item has a unique simulator-derived target answer; the agent must write a PHREEQC input file, execute the simulator (PHREEQC 3.8.6 with an author-supplied validated \texttt{phreeqc.dat} thermodynamic database), inspect generated output files, and commit to its final answer by writing a file containing only the letter (A, B, C, or D); missing or unparseable final-answer files count as incorrect. We balanced the truth-label distribution to be approximately uniform (A=\BenchAlwaysA, B=\BenchAlwaysB, C=\BenchAlwaysC, D=\BenchAlwaysD); even so, LLMs exhibit answer-label preferences under one-shot prompting (e.g., Claude Sonnet 4.6 prefers C, GPT-5.4 prefers B; Appendix~\ref{sec:appendix-label-behavior}). For the scenario-clustered analysis (Appendix~\ref{sec:appendix-scenario-stratification}), a scenario stem is operationally defined as the first paragraph of the problem statement after whitespace normalization.

\subsection{Agent Harness}
\label{sec:harness}

\begin{figure}[t]
\centering
\includegraphics[width=0.85\linewidth]{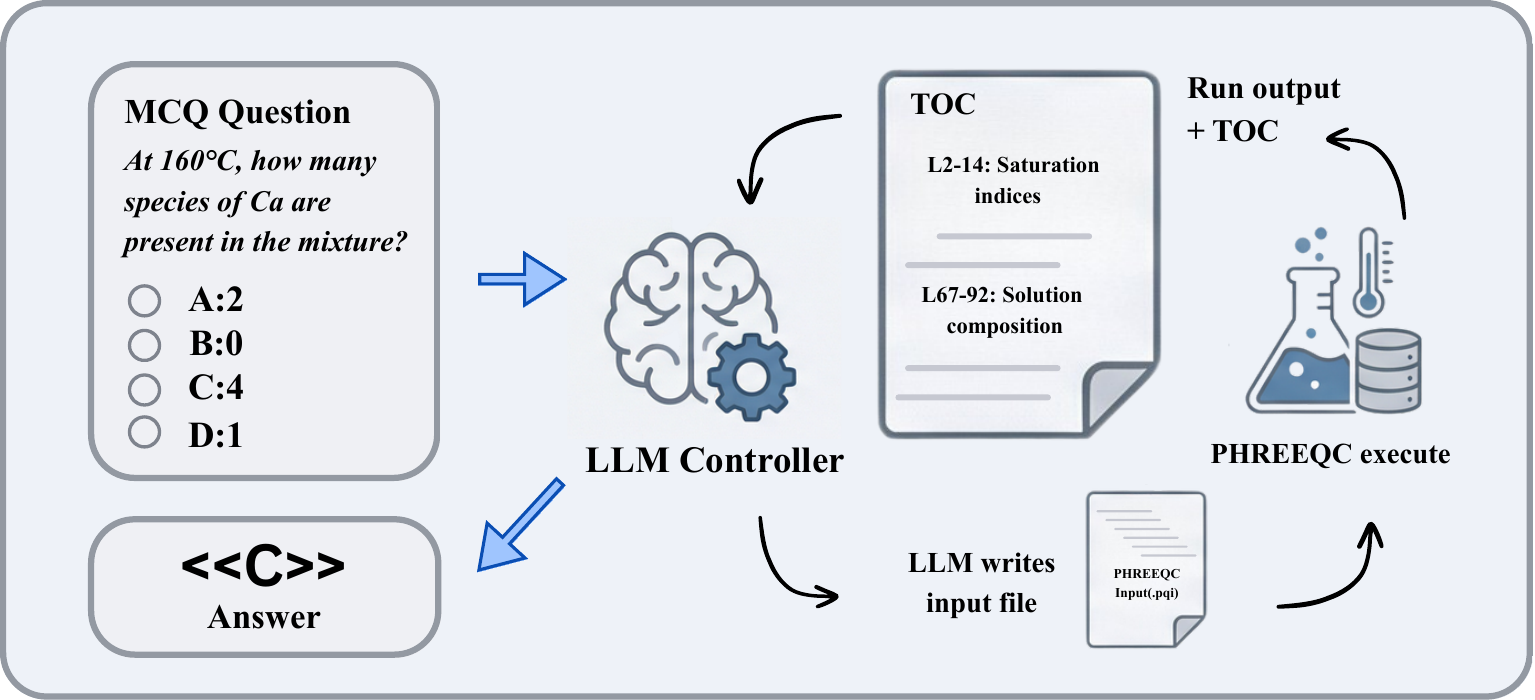}
\caption{\BenchName{} agent harness under the Agent-TOC protocol.}
\label{fig:agent-workflow}
\end{figure}

The agent workflow is shown in Figure~\ref{fig:agent-workflow}. The custom agent is given four tools --- \texttt{write\_file}, \texttt{read\_file}, \texttt{list\_file}, and \texttt{execute\_phreeqc} --- and runs each question in an isolated workspace with a fixed maximum of 24 agent steps. PHREEQC execution is deterministic for a given input file and the frozen benchmark database; outputs are natively section-structured (Solution composition, Distribution of species, Saturation indices, $\dots$) and can reach \OutputMaxChars{}, motivating a structured-retrieval interface rather than always inlining the full payload. We evaluate two output-access protocols that differ only in what \texttt{execute\_phreeqc} returns. Under \textbf{Agent-TOC}, the tool writes the simulator output to a workspace file and returns a section-level outline (section names paired with starting line numbers) so the agent can fetch evidence on demand via \texttt{read\_file} with explicit line ranges; the size threshold is set to \texttt{MAX\_FULL\_CHARS=10M} characters (further details in Appendices~\ref{sec:appendix-toc-impl} and~\ref{sec:appendix-cap-affected-per-model}). Under \textbf{Agent-Raw100k}, the output is inlined head/tail-truncated to \RawCapChars{} characters.

\section{Main Results}
\label{sec:main-results}

Table~\ref{tab:main-results} reports direct, chain-of-thought (CoT), TOC-agent, and Raw100k-agent accuracy across model families on \BenchName. Direct baselines sit in the \GPTFiveFourOneShot--\OpusOneShot{} range, just above the \BenchMajority{} majority-label floor; without tool access, models exhibit strong answer-position preferences (Appendix~\ref{sec:appendix-label-behavior}). All percentages use the \BenchN-item denominator with ungraded answers counted as incorrect. We read the table along three axes below: whether mental reasoning can substitute for tool grounding (§\ref{sec:main-cot}), what tool access adds for capable models (§\ref{sec:main-tool}), and when tool augmentation backfires (§\ref{sec:main-belowmid}).

\begin{table}[!ht]
\centering
\footnotesize
\setlength{\tabcolsep}{4pt}
\caption{Main results on \BenchName. All percentages use all \BenchN{} items as the denominator; unparseable or missing final answers count as incorrect. CIs are 95\% Wilson binomial. Raw100k denotes the truncated raw-output agent condition. Capability-tier groupings are informal and motivated by the TOC-vs-Raw pattern in Section~\ref{sec:diagnostic}.}
\label{tab:main-results}
\begin{tabular}{llcccc}
\toprule
Tier & Model & Direct (95\% CI) & CoT (95\% CI) & Agent-TOC & Agent-Raw100k \\
\midrule
\multirow{3}{*}{Top} & Claude Opus 4.6 & \OpusOneShot~\OpusOneShotCI & \OpusCoT~\OpusCoTCI & \OpusTOC & \OpusRaw \\
 & Claude Sonnet 4.6 & \SonnetOneShot~\SonnetOneShotCI & \SonnetCoT~\SonnetCoTCI & \SonnetTOC & \SonnetRaw \\
 & GPT-5.4 & \GPTFiveFourOneShot~\GPTFiveFourOneShotCI & \GPTFiveFourCoT~\GPTFiveFourCoTCI & \GPTFiveFourTOC & \GPTFiveFourRaw \\
\midrule
\multirow{2}{*}{Mid} & GPT-5.2 & \GPTFiveTwoOneShot~\GPTFiveTwoOneShotCI & \GPTFiveTwoCoT~\GPTFiveTwoCoTCI & \GPTFiveTwoTOC & \GPTFiveTwoRaw \\
 & GPT-5.1 & \GPTFiveOneOneShot~\GPTFiveOneOneShotCI & \GPTFiveOneCoT~\GPTFiveOneCoTCI & \GPTFiveOneTOC & \GPTFiveOneRaw \\
\midrule
Below-mid & Gemini 3 Flash Preview & \GeminiThreeFlashOneShot~\GeminiThreeFlashOneShotCI & \GeminiThreeFlashCoT~\GeminiThreeFlashCoTCI & \GeminiThreeFlashTOC & \GeminiThreeFlashRaw \\
\bottomrule
\end{tabular}
\end{table}

\subsection{Chain-of-Thought Is Not Sufficient}
\label{sec:main-cot}

A geochemistry question may be solvable through mental recall and arithmetic, or it may require numerical computation that no amount of pre-trained knowledge can deliver to the precision the four answer choices demand. We probe this with chain-of-thought (CoT) prompting~\citep{wei2022chainofthought}, which has been shown to substantially boost performance on reasoning-heavy tasks. Across model families, CoT yields roughly $\pm 8$\,pp around direct (Opus and GPT-5.1 lose 4.0\,pp; Sonnet, GPT-5.4, and GPT-5.2 gain 6.0--8.0\,pp). The contrast against the output-token expansion CoT produces is sharp: Sonnet expands its total output tokens by 140$\times$ (1.6K to 225K) for a $+6.0$\,pp accuracy gain, and GPT-5.1 expands by 29$\times$ (2.0K to 58.6K) for a $-4.0$\,pp \emph{decrease} (Appendix~\ref{sec:appendix-cot-tokens}). For items not solvable from parametric knowledge, scaling the reasoning-token budget cannot substitute for grounded simulator output: the bottleneck on \BenchName{} is computational, not reasoning-budgeted.

\subsection{Tool Access Is Necessary}
\label{sec:main-tool}

Tool access (Agent-TOC) adds 15.0--41.5\,pp on top of the better no-tool baseline for the five top- and mid-tier completed agents: Opus $+41.5$\,pp, Sonnet $+41.0$\,pp, GPT-5.4 $+37.5$\,pp, GPT-5.2 $+19.5$\,pp, GPT-5.1 $+15.0$\,pp. Both Anthropic and OpenAI top-tier models exhibit similar gains, supporting the reading that simulator access --- not vendor-specific scaffolding --- drives the improvement. The headline gains in Table~\ref{tab:main-results} are attributable to PHREEQC computation, not to additional reasoning tokens: when a question exceeds what a model can compute mentally, a calculator is necessary.

\subsection{Tool Augmentation Can Hurt Below-Mid-Tier Models}
\label{sec:main-belowmid}

Gemini 3 Flash Preview's TOC agent reverses the pattern, reaching only \GeminiThreeFlashTOC{} --- a \GeminiThreeFlashTOCMinusBestNoTool{} regression below its better no-tool baseline (CoT, \GeminiThreeFlashCoT{}) and the only completed agent in our evaluation where tool access actively harms aggregate performance. The mechanism is step-budget exhaustion under TOC navigation overhead: switching to Raw100k recovers headline accuracy to \GeminiThreeFlashRaw{}, with detailed trajectory analysis in Appendix~\ref{sec:appendix-gemini-step-budget}. We read this as a third capability tier --- below-mid --- where tool augmentation can degrade aggregate accuracy not by producing wrong answers but by preventing the model from producing any answer at all. Even Gemini Flash 3's recovered Raw100k accuracy (\GeminiThreeFlashRaw{}, $+8.5$\,pp over Direct) is roughly half the smallest Direct-to-Tool gain achieved by the five top- and mid-tier agents (GPT-5.1: $+15.0$\,pp; Section~\ref{sec:main-tool}), so even when navigation overhead is removed, tool access helps Gemini Flash 3 less than any other model in our evaluation.

\section{Beyond Aggregate Accuracy: Diagnostic Probes for Tool-Augmented Agents}
\label{sec:diagnostic}

Section~\ref{sec:main-results} establishes that tool access drives the headline gains for top- and mid-tier models and that it actively harms the below-mid tier. This section opens the box behind those aggregate numbers along two axes: whether tool use is pure addition or trades items at the unit level (§\ref{sec:diag-retention}), and whether the gains depend on the choice of output-access protocol (§\ref{sec:toc}). Three additional diagnostics --- a matched 50-question external-harness comparison against Claude Code CLI and OpenAI Codex CLI showing 2.7--3.3$\times$ better tokens-per-correct for the custom harness (Appendix~\ref{sec:appendix-external-harness}), trajectory and operational cost (Appendix~\ref{sec:appendix-trajectory}), and scenario-clustered uncertainty (Appendix~\ref{sec:appendix-scenario-stratification}) --- support the same capability-tier reading.

\subsection{Retention: Tool Use Trades Items, Not Just Adds Them}
\label{sec:diag-retention}

Aggregate accuracy alone cannot tell whether tool access reliably improves a model. An agent can solve new questions through grounded computation while also losing questions that the same model answered correctly without tools. Table~\ref{tab:retention} therefore decomposes each agent run into kept, gained, and lost answers relative to its Direct no-tool baseline, reporting both output-access protocols (Agent-TOC and Agent-Raw100k) side by side. Retention is kept divided by kept plus lost. Higher retention indicates greater stability under tool augmentation: among questions the model answered correctly without tools, fewer are lost after adding the PHREEQC agent loop. The rightmost column reports the TOC$-$Raw100k retention difference, separating per-item navigation cost specific to the TOC interface from any shared capability gap visible under both protocols.

\begin{table}[t]
\centering
\footnotesize
\setlength{\tabcolsep}{3.5pt}
\caption{Gain/loss/retention decomposition relative to each model's Direct no-tool baseline, with Agent-TOC and Agent-Raw100k reported side by side. Retention $=$ kept/(kept+lost). $\Delta$~Ret.\ and $\Delta$~Net (TOC minus Raw100k) isolate the per-item cost of the TOC navigation interface from any shared capability gap visible under both protocols.}
\label{tab:retention}
\begin{tabular}{lcccccccccc}
\toprule
& \multicolumn{4}{c}{Direct $\rightarrow$ Agent-TOC} & \multicolumn{4}{c}{Direct $\rightarrow$ Agent-Raw100k} & \multicolumn{2}{c}{TOC vs Raw100k} \\
\cmidrule(lr){2-5} \cmidrule(lr){6-9} \cmidrule(lr){10-11}
Model & Kept & Gain & Lost & Ret. & Kept & Gain & Lost & Ret. & $\Delta$ Ret. & $\Delta$ Net \\
\midrule
Claude Opus 4.6   & \OpusKept           & \OpusGained           & \OpusLost           & \OpusRetention            & \OpusRawKept         & \OpusRawGained       & \OpusRawLost       & \OpusRawRetention         & \OpusTOCMinusRawRetention      & $+1$ \\
Claude Sonnet 4.6 & \SonnetKept         & \SonnetGained         & \SonnetLost         & \SonnetRetention          & \SonnetRawKept       & \SonnetRawGained     & \SonnetRawLost     & \SonnetRawRetention       & \SonnetTOCMinusRawRetention    & $+1$ \\
GPT-5.4           & \GPTFiveFourKept    & \GPTFiveFourGained    & \GPTFiveFourLost    & \GPTFiveFourRetention     & \GPTFiveFourRawKept  & \GPTFiveFourRawGained & \GPTFiveFourRawLost & \GPTFiveFourRawRetention  & \GPTFiveFourTOCMinusRawRetention  & $+5$ \\
GPT-5.2           & \GPTFiveTwoKept     & \GPTFiveTwoGained     & \GPTFiveTwoLost     & \GPTFiveTwoRetention      & \GPTFiveTwoRawKept   & \GPTFiveTwoRawGained  & \GPTFiveTwoRawLost  & \GPTFiveTwoRawRetention   & \GPTFiveTwoTOCMinusRawRetention   & $-15$ \\
GPT-5.1           & \GPTFiveOneKept     & \GPTFiveOneGained     & \GPTFiveOneLost     & \GPTFiveOneRetention      & \GPTFiveOneRawKept   & \GPTFiveOneRawGained  & \GPTFiveOneRawLost  & \GPTFiveOneRawRetention   & \GPTFiveOneTOCMinusRawRetention   & $-19$ \\
\bottomrule
\end{tabular}
\end{table}

Two patterns generalize across the five agents in Table~\ref{tab:retention} without requiring item-level recitation. First, the GPT family is less stable under tool augmentation than the Anthropic family: according to Appendix~\ref{sec:appendix-lost-by-bin}, GPT-5.2 is the least stable, losing 43.6\% of its direct-correct items overall versus 13.5\% for Sonnet and 16.7\% for Opus ($\sim$3$\times$ the Anthropic rate). The gap persists even on the smallest-output bin ($\le$10k characters, where all five agents face the same evidence load): GPT-5.2 still loses 50\% of the items it answered correctly without tools, roughly 5$\times$ Sonnet's 9.1\% on the same bin (Appendix~\ref{sec:appendix-lost-by-bin}). Second, different models lose different items: the lost-item sets only partially overlap across backbones, which rules out a ``broken benchmark'' explanation in which a fixed set of items is unsolvable under tool use. The regressions instead trace to each model's specific tool-loop behavior. Combined with the Gemini non-commit failure in Section~\ref{sec:main-belowmid}, this establishes a benchmark-level claim that aggregate accuracy cannot make: tool augmentation is never purely additive --- top- and mid-tier models trade items even when net-positive, and the below-mid model fails to commit at all.

\subsection{TOC Pays Off Only for Top-Tier Models}
\label{sec:toc}

\begin{figure}[t]
\centering
\includegraphics[width=\linewidth]{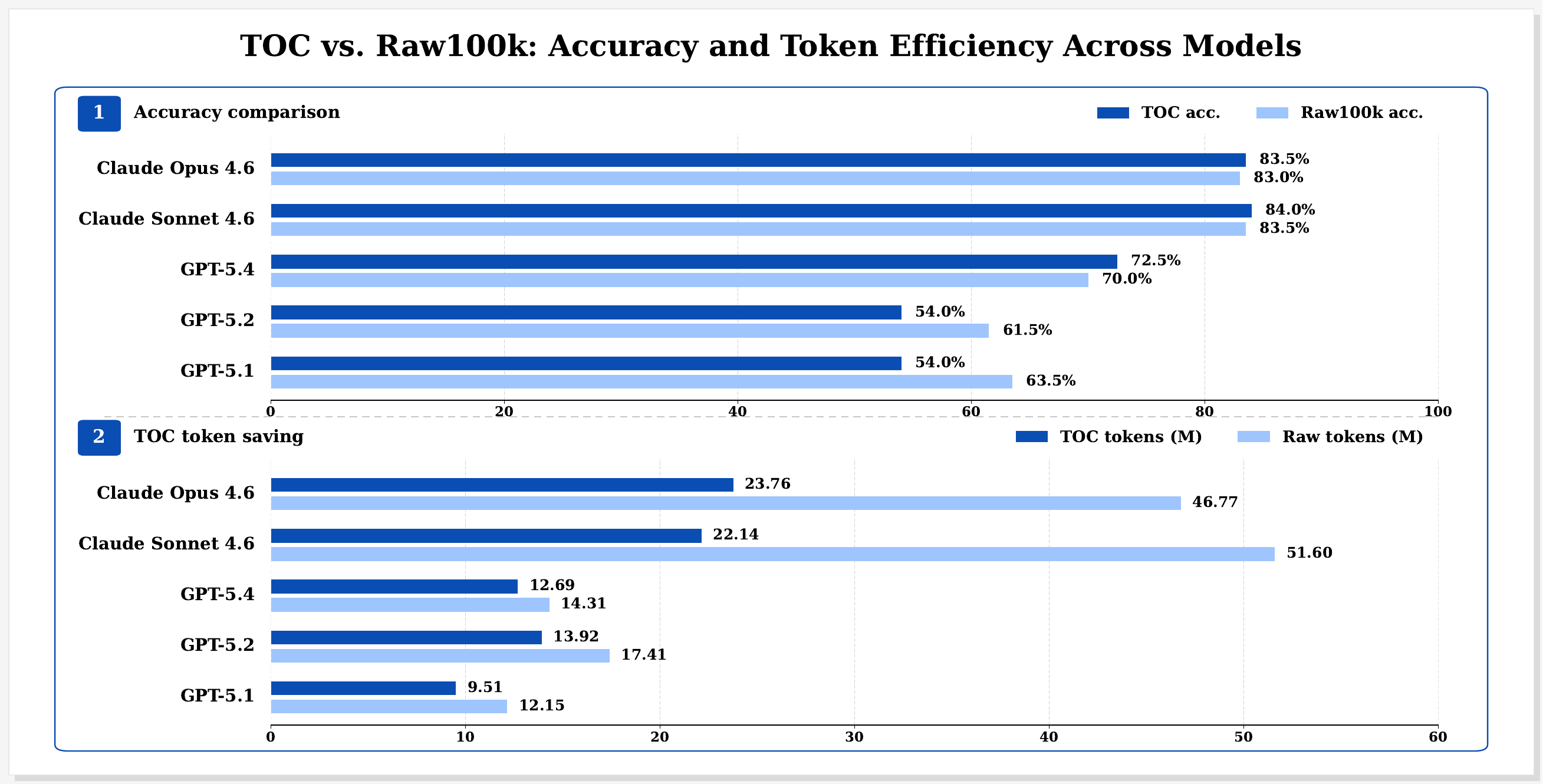}
\caption{TOC versus truncated raw-output execution interface, per model. Both protocols use the same agent loop and differ only in what \texttt{execute\_phreeqc} returns; raw observations are head/tail-truncated to \RawCapChars{} characters per PHREEQC execution. Each point summarizes one model's full-benchmark TOC and Raw100k accuracy and total agent input-token cost (millions).}
\label{fig:toc-raw}
\end{figure}

TOC keeps raw output out of context at the cost of an extra navigation step the agent must execute. Whether that tradeoff pays off should depend on the model's navigation capability, not on any property of the benchmark. Figure~\ref{fig:toc-raw} compares the TOC and Raw100k protocols (Section~\ref{sec:harness}) on the full benchmark, where the \RawCapChars-character truncation is non-binding for most items and the question is whether structured access changes cost or accuracy.

Figure~\ref{fig:toc-raw} shows two patterns. The top tier (Opus, Sonnet, GPT-5.4) matches or exceeds raw-output accuracy by 0.5--2.5\,pp at 11--57\% input-token reductions; TOC is Pareto-good in this evaluation. The mid tier (GPT-5.1, GPT-5.2) regresses by 7.5--9.5\,pp at similar token savings. The cut crosses vendor: GPT-5.4 sits with the Anthropic top tier, while same-vendor GPT-5.1/5.2 sit on the mid-tier side. The mid-tier gap is not a rerun artifact: \TOCRerunNumSamples{} independent TOC reruns per model in Table~\ref{tab:toc-rerun-variance} (all five top- and mid-tier completed agents) yield top-tier accuracy SDs of \SonnetTOCRerunSD, \OpusTOCRerunSD, \GPTFiveFourTOCRerunSD\,pp (Sonnet, Opus, GPT-5.4) and a maximum SD of \TOCRerunMaxSD\,pp on GPT-5.1, all smaller than the 7.5--9.5\,pp gap. Token consumption is even more stable in relative terms --- post-splice rerun token SD ranges from 0.12\,M (GPT-5.1) to 1.77\,M (Opus), at most 7.8\% of mean per-rerun input tokens --- so the 11--57\% TOC token-reduction range reflects a property of the protocol rather than rerun noise.

\begin{table}[t]
\centering
\footnotesize
\setlength{\tabcolsep}{3pt}
\caption{TOC rerun stability across all five top- and mid-tier agents on \BenchName{}. $\Delta$ columns are reruns minus original (Figure~\ref{fig:toc-raw}); tokens are total agent input tokens (M). All TOC values reflect the spliced configuration in which the 45-item uncapped-TOC fix is applied to each rerun's long-output items, matching the headline accuracy. Per-item decomposition for GPT-5.1 in Appendix~\ref{sec:appendix-stability}.}
\label{tab:toc-rerun-variance}
\begin{tabular}{lccccc}
\toprule
Model & Orig.\ (acc, tok) & Acc $\Delta$ (pp) & Tok $\Delta$ (M) & Acc SD / Rng & Tok SD / Rng \\
\midrule
Claude Opus 4.6 + TOC   & \OpusTOC, \OpusTOCTokens             & \OpusTOCRerunAccDiff       & \OpusTOCRerunTokDiff       & \OpusTOCRerunSD{} / \OpusTOCRerunRange       & \OpusTOCRerunTokSD{} / \OpusTOCRerunTokRange \\
Claude Sonnet 4.6 + TOC & \SonnetTOC, \SonnetTOCTokens & \SonnetTOCRerunAccDiff & \SonnetTOCRerunTokDiff & \SonnetTOCRerunSD{} / \SonnetTOCRerunRange & \SonnetTOCRerunTokSD{} / \SonnetTOCRerunTokRange \\
GPT-5.1 + TOC & \GPTFiveOneTOC, \GPTFiveOneTOCTokens & \GPTFiveOneTOCRerunAccDiff & \GPTFiveOneTOCRerunTokDiff & \GPTFiveOneTOCRerunSD{} / \GPTFiveOneTOCRerunRange & \GPTFiveOneTOCRerunTokSD{} / \GPTFiveOneTOCRerunTokRange \\
GPT-5.2 + TOC & \GPTFiveTwoTOC, \GPTFiveTwoTOCTokens & \GPTFiveTwoTOCRerunAccDiff & \GPTFiveTwoTOCRerunTokDiff & \GPTFiveTwoTOCRerunSD{} / \GPTFiveTwoTOCRerunRange & \GPTFiveTwoTOCRerunTokSD{} / \GPTFiveTwoTOCRerunTokRange \\
GPT-5.4 + TOC & \GPTFiveFourTOC, \GPTFiveFourTOCTokens & \GPTFiveFourTOCRerunAccDiff & \GPTFiveFourTOCRerunTokDiff & \GPTFiveFourTOCRerunSD{} / \GPTFiveFourTOCRerunRange & \GPTFiveFourTOCRerunTokSD{} / \GPTFiveFourTOCRerunTokRange \\
\bottomrule
\end{tabular}
\end{table}

\paragraph{Are the TOC savings biased toward long-output items?}
If TOC's full-benchmark token reduction (Figure~\ref{fig:toc-raw}) came entirely from items where Raw100k inlines a 100k payload, the small-output bulk could be paying TOC's structured-retrieval overhead with no benefit and the long tail could carry all the gains. To rule this out we split the benchmark into the \RawCapUnaffectedN{}-item small-output bulk (output $\le$100k in the canonical reference run) and the 45-item long-output tail (items whose simulator outputs exceed 100k characters in at least one of the five completed TOC runs; Appendix~\ref{sec:appendix-cap-affected-per-model}) and report TOC versus Raw100k on each slice in Table~\ref{tab:capaffected-toc-verification}. We use the 5-model union (45) rather than the canonical 37 ($=$ 200 $-$ \RawCapUnaffectedN{}) for the long-output tail because per-model PHREEQC inputs differ: an item small-output for one model can be long-output for another, so the union covers every (model, item) pair that could plausibly hit the 100k regime. Rerun stability of TOC accuracy on the small-output-bulk slice itself is reported in Appendix~\ref{sec:appendix-stability-small} (max small-bulk SD \GPTFiveOneTOCSmallRerunSD\,pp on GPT-5.1, top-tier SDs \SonnetTOCSmallRerunSD--\GPTFiveFourTOCSmallRerunSD\,pp).

\begin{table}[t]
\centering
\footnotesize
\setlength{\tabcolsep}{4pt}
\caption{Small-output bulk uses the \RawCapUnaffectedN{} items with $\le$100k characters in the canonical reference run; long-output tail uses the 5-model union of items exceeding 100k in at least one TOC run (Appendix~\ref{sec:appendix-cap-affected-per-model}). We use 45 rather than 37 ($=$ 200 $-$ \RawCapUnaffectedN{}) because per-model PHREEQC inputs differ --- an item small-output for one model can be long-output for another, so the union covers every (model, item) pair that could plausibly hit the 100k regime; some items therefore appear under both row groups for at least one model.}
\label{tab:capaffected-toc-verification}
\begin{tabular}{lccccccc}
\toprule
Model & TOC acc. & Raw100k acc. & $\Delta$ acc. & TOC tok/q & Raw100k tok/q & TOC token saving \\
\midrule
\multicolumn{7}{l}{\textit{Small-output bulk (\RawCapUnaffectedN{} items, $\le$100k characters in canonical reference)}} \\
\midrule
Claude Opus 4.6   & \OpusTOCCapUnaffected            & \OpusRawCapUnaffected        & \OpusTOCMinusRawCapUnaffected           & $\sim$98{,}000 & 181{,}024 & $\sim$46\% \\
Claude Sonnet 4.6 & \SonnetTOCCapUnaffected          & \SonnetRawCapUnaffected      & \SonnetTOCMinusRawCapUnaffected         & 82{,}648  & 185{,}059 & 55\% \\
GPT-5.4           & \GPTFiveFourTOCCapUnaffected     & \GPTFiveFourRawCapUnaffected & \GPTFiveFourTOCMinusRawCapUnaffected     & 54{,}401  & 52{,}514  & $-$3.6\% \\
GPT-5.2           & \GPTFiveTwoTOCCapUnaffected      & \GPTFiveTwoRawCapUnaffected  & \GPTFiveTwoTOCMinusRawCapUnaffected      & 54{,}792  & 57{,}108  & 4\% \\
GPT-5.1           & \GPTFiveOneTOCCapUnaffected      & \GPTFiveOneRawCapUnaffected  & \GPTFiveOneTOCMinusRawCapUnaffected      & 30{,}426  & 36{,}604  & 17\% \\
\midrule
\multicolumn{7}{l}{\textit{Long-output tail (45 items, $>$100k characters in at least one TOC run)}} \\
\midrule
Claude Opus 4.6   & \OpusTOCFortyFiveAcc        & \OpusRawFortyFiveAcc         & \OpusTOCMinusRawFortyFive        & \OpusTOCFortyFiveTokensPerItem        & \OpusRawFortyFiveTokensPerItem        & 56\% \\
Claude Sonnet 4.6 & \SonnetTOCFortyFiveAcc      & \SonnetRawFortyFiveAcc       & \SonnetTOCMinusRawFortyFive      & \SonnetTOCFortyFiveTokensPerItem      & \SonnetRawFortyFiveTokensPerItem      & 65\% \\
GPT-5.4           & \GPTFiveFourTOCFortyFiveAcc & \GPTFiveFourRawFortyFiveAcc  & \GPTFiveFourTOCMinusRawFortyFive & \GPTFiveFourTOCFortyFiveTokensPerItem & \GPTFiveFourRawFortyFiveTokensPerItem & 42\% \\
GPT-5.2           & \GPTFiveTwoTOCFortyFiveAcc  & \GPTFiveTwoRawFortyFiveAcc   & \GPTFiveTwoTOCMinusRawFortyFive  & \GPTFiveTwoTOCFortyFiveTokensPerItem  & \GPTFiveTwoRawFortyFiveTokensPerItem  & 40\% \\
GPT-5.1           & \GPTFiveOneTOCFortyFiveAcc  & \GPTFiveOneRawFortyFiveAcc   & \GPTFiveOneTOCMinusRawFortyFive  & \GPTFiveOneTOCFortyFiveTokensPerItem  & \GPTFiveOneRawFortyFiveTokensPerItem  & 7\% \\
\bottomrule
\end{tabular}
\end{table}

Both slices show the same top-vs-mid-tier ordering. On the small-output bulk, top-tier gains accuracy under TOC: Opus $+$1.2\,pp at $\sim$46\% token saving (Pareto-win), Sonnet $+$1.8\,pp at 55\% token saving (Pareto-win), GPT-5.4 $+$4.9\,pp at 3.6\% token overhead; mid-tier (GPT-5.1, GPT-5.2) lose 7.4--12.9\,pp. On the long-output tail, top-tier Opus, Sonnet, and GPT-5.4 trail Raw100k by 2--5\,pp; GPT-5.2 trails by 8.9\,pp; GPT-5.1 reverses to win by 6.7\,pp because Raw100k truncates most heavily on these items. TOC saves 7--65\% of input tokens on the long-output tail. For top-tier models, TOC's long-tail token saving (Opus 56\%, Sonnet 65\%, GPT-5.4 42\%) is substantially larger than the same model's small-bulk saving (46\%, 55\%, $-$3.6\% overhead) --- the 2--5\,pp accuracy give-back buys disproportionately large token reductions on exactly the items where Raw100k is most expensive. Per-output-size bin breakdown is in Appendix~\ref{sec:appendix-output-bin-grid}.

A natural alternative to TOC on this long-tail slice is to simply enlarge the inline truncation window rather than introduce structured retrieval. Table~\ref{tab:large-output-access} compares Raw100k and Raw500k on the \LargeOutputN{} long-output items (GPT family only; Raw500k was not run for Anthropic models). Across all three GPT models the larger window does not consistently improve accuracy and substantially raises per-question token cost, so TOC's structured retrieval is not interchangeable with a simple window enlargement.

\begin{table}[t]
\centering
\footnotesize
\setlength{\tabcolsep}{4pt}
\caption{Raw100k versus Raw500k on the same \LargeOutputN{} long-output items as Table~\ref{tab:capaffected-toc-verification} (GPT family only; Raw500k was not run for Anthropic models). Across all three GPT models, the larger inline window does not consistently improve accuracy and substantially raises per-question token cost. Token columns report mean input tokens per question on this subset.}
\label{tab:large-output-access}
\begin{tabular}{lccccc}
\toprule
Model & Window & Raw100k acc. & Raw500k acc. & Raw100k tok/q & Raw500k tok/q \\
\midrule
GPT-5.4 & \GPTFiveFourContextWindow & \GPTFiveFourRawLargeAcc & \GPTFiveFourRawFiveHundredKLargeAcc & \GPTFiveFourRawLargeTokensPerItem & \GPTFiveFourRawFiveHundredKLargeTokensPerItem \\
GPT-5.2 & \GPTFiveTwoContextWindow & \GPTFiveTwoRawLargeAcc & \GPTFiveTwoRawFiveHundredKLargeAcc & \GPTFiveTwoRawLargeTokensPerItem & \GPTFiveTwoRawFiveHundredKLargeTokensPerItem \\
GPT-5.1 & \GPTFiveOneContextWindow & \GPTFiveOneRawLargeAcc & \GPTFiveOneRawFiveHundredKLargeAcc & \GPTFiveOneRawLargeTokensPerItem & \GPTFiveOneRawFiveHundredKLargeTokensPerItem \\
\bottomrule
\end{tabular}
\end{table}

\section{Failure Analysis}
\label{sec:failure}

We complement the automatic log-derived categorization in Appendix~\ref{sec:appendix-trajectory} with a domain-expert review of the residual failure cases on \BenchName{}. The review surfaces a capability-tier $\times$ failure-mode asymmetry that aggregate accuracy hides: mid-tier GPT failures are dominated by high-confidence wrong PHREEQC builds together with cases where the right answer is present in \texttt{result.out} but the agent does not use it, while top-tier Opus failures concentrate in subtle bad-PHREEQC-input cases (the input parses and runs but contains a modeling flaw such as an omitted \texttt{charge} keyword or a stiff kinetic constant) and PHREEQC retry loops that exhaust the step budget without committing. Tool-loop failures therefore concentrate in different regions of the (input, execute, navigate, commit) chain by capability tier. Full categorical taxonomy and per-category descriptions are in Appendix~\ref{sec:appendix-failure-taxonomy}.

\section{Discussion}

\paragraph{Output-access protocol $\times$ model capability is an interaction effect.}
TOC's payoff tracks capability, not vendor: GPT-5.4 behaves like the Anthropic top-tier (TOC Pareto-good at substantial token savings), while same-vendor GPT-5.1 and GPT-5.2 lose 7.5--9.5\,pp under TOC at marginal token savings (Section~\ref{sec:toc}). Output-access protocol should be matched to the deployed model's capability rather than chosen uniformly across a model fleet, and output-access ablation should be reported alongside aggregate accuracy whenever a benchmark involves long simulator outputs.

\paragraph{TOC token saving across simulator domains.}
The structured-retrieval mechanism is mechanical: any deterministic simulator with section-structured output can build a TOC-style handle, replacing an inline payload that grows with output size. The capability-tier interaction we report likely transfers under the same mechanism: top-tier models that navigate structure successfully reap the savings; mid-tier models that fail navigation lose accuracy. What does not transfer cleanly is the model-specific reasoning-token expansion under inline content we observed (Section~\ref{sec:toc}) --- that is a property of how a particular model expands reasoning when given inline simulator output, not of the simulator --- so cross-simulator validation of TOC's per-model token effect would need separate measurement per (simulator, model) pair.

\paragraph{Rerun stability and retention are independent properties.}
GPT-5.2 has the lowest retention (\GPTFiveTwoRetention) but a small TOC rerun SD of \GPTFiveTwoTOCRerunSD\,pp (close to Sonnet's \SonnetTOCRerunSD\,pp); GPT-5.1 has higher retention (\GPTFiveOneRetention) but the largest rerun SD (\GPTFiveOneTOCRerunSD\,pp; Appendix~\ref{sec:appendix-stability}). Most benchmark papers conflate these as ``noise''; reporting them separately exposes structural agent-protocol properties that aggregate accuracy hides, and we recommend both in future tool-augmented agent evaluations.

\paragraph{Long-output set size is itself a capability signal.}
On a simulator like PHREEQC the set of items that produce long outputs is not a dataset constant — it depends on how completely the agent's PHREEQC input exercises the simulator. Across our five completed TOC runs the per-model count of items with output exceeding 100k characters ranges from 23 (GPT-5.1) to 41 (GPT-5.4), with union 45 and intersection only 15 (Appendix~\ref{sec:appendix-cap-affected-per-model}). Stronger agents both produce more long outputs (more complete inputs) and navigate them better, so long-output coverage and long-output accuracy compound rather than trade off. Future tool-augmented benchmark designers should report per-model output-size statistics rather than treating output size as an item-level constant.

\section{Limitations}

\paragraph{MCQ format and benchmark scope.}
Multiple-choice grading enables controlled exact-match scoring on \BenchN{} items derived from \BenchScenarioN{} scenario stems with a fixed author-supplied PHREEQC thermodynamic database. We report scenario-level metrics (Appendix~\ref{sec:appendix-scenario-stratification}) so items are not treated as fully independent scientific setups; results should be interpreted as measurements under this benchmark format rather than as unconstrained scientific modeling performance.

\section{Conclusion}

We introduced \BenchName, a benchmark for evaluating LLM agents that operate deterministic scientific simulators under a unique-answer setting. The benchmark reframes PHREEQC from a narrow domain application into a controlled testbed for studying tool-augmented scientific computation. The central findings, organized around benchmark diagnosis rather than geochemistry, are: tool access produces net gains across the five top- and mid-tier completed agents at variable retention costs (56.4--86.5\%) but regresses below the no-tool baseline on the below-mid Gemini Flash 3 agent; chain-of-thought does not substitute for grounded execution; output-access protocol effects align with model capability tier rather than vendor, with a metadata-guided interface (TOC) acting as a Pareto-good interface for top-tier models and a 7.5--9.5\,pp accuracy cost for mid-tier models; rerun stability and item-level retention are independent properties of an agent and should be reported separately; and failure mechanism is itself capability-tier-dependent, with top-tier failures dominated by step-budget exhaustion after committing, mid-tier by premature commitment, and below-mid by step-budget exhaustion without committing. The result is a reusable diagnostic harness for measuring scientific tool-use agents and locating where reliability is lost.

\bibliographystyle{plainnat}
\bibliography{references}

\newpage
\appendix

\section{Answer-Label Behavior}
\label{sec:appendix-label-behavior}

The MCQ surface can be distorted by label imbalance or model answer-position preferences. As a robustness check, Table~\ref{tab:label-behavior} reports two metrics per method: (i)~the \emph{predicted-label distribution} --- how often the model picked each of A/B/C/D across the 200 items --- and (ii)~the \emph{per-true-label accuracy range} --- the lowest and highest accuracy across the four true-label classes. The benchmark's true-label distribution is A/B/C/D = \BenchTruthCounts, so an unbiased model would predict each label in roughly the same proportions; a model that picks one label disproportionately is biased.

Direct no-tool baselines exhibit strong label preferences. Sonnet direct predicts C \emph{101 times} against only 48 true-C items, scoring 68.8\% on true-C but only 12.8\% on true-B. GPT-5.4 direct over-predicts B (109/200 vs 47 true-B), reaching just 7.9\% on true-A. GPT-5.2 direct shows the same B preference (101/200), and GPT-5.1 direct is closer to balanced but still skews toward B (62/200). The Anthropic + TOC agents flatten the predicted distribution toward the true-label counts and tighten the per-true-label accuracy range upward to spans of $\le$5\,pp (Opus + TOC: \OpusTOCLabelRange; Sonnet + TOC: \SonnetTOCLabelRange). The GPT family under TOC retains wider per-label spreads (GPT-5.4: \GPTFiveFourTOCLabelRange{}, span 18\,pp; GPT-5.2: \GPTFiveTwoTOCLabelRange{}, span 26\,pp; GPT-5.1: \GPTFiveOneTOCLabelRange{}, span 27\,pp), and Gemini Flash 3 + TOC fails to commit on 150 of 200 items entirely (\GeminiThreeFlashTOCPredDist; Section~\ref{sec:main-belowmid}). The label-behavior table thus supports two readings used elsewhere in the paper: it is the empirical evidence cited in Section~\ref{sec:benchmark} that direct accuracies sit near the majority-label floor by construction rather than by mechanical guessing, and the Anthropic-versus-GPT spread gap reinforces the GPT-family-instability pattern observed in Sections~\ref{sec:diag-retention} and~\ref{sec:toc}.

\begin{table}[h]
\centering
\small
\caption{Answer-label behavior for selected completed runs. Direct baselines can show strong label preferences, whereas the strongest TOC agents predict labels close to the true-label distribution.}
\label{tab:label-behavior}
\begin{tabular}{lcc}
\toprule
Method & Predicted A/B/C/D/NA & Per-true-label accuracy range \\
\midrule
Claude Opus 4.6 direct & \OpusDirectPredDist & \OpusDirectLabelRange \\
Claude Sonnet 4.6 direct & \SonnetDirectPredDist & \SonnetDirectLabelRange \\
GPT-5.4 direct & \GPTFiveFourDirectPredDist & \GPTFiveFourDirectLabelRange \\
GPT-5.2 direct & \GPTFiveTwoDirectPredDist & \GPTFiveTwoDirectLabelRange \\
GPT-5.1 direct & \GPTFiveOneDirectPredDist & \GPTFiveOneDirectLabelRange \\
Gemini 3 Flash Preview direct & \GeminiThreeFlashDirectPredDist & \GeminiThreeFlashDirectLabelRange \\
\midrule
Claude Opus 4.6 + TOC & \OpusTOCPredDist & \OpusTOCLabelRange \\
Claude Sonnet 4.6 + TOC & \SonnetTOCPredDist & \SonnetTOCLabelRange \\
GPT-5.4 + TOC & \GPTFiveFourTOCPredDist & \GPTFiveFourTOCLabelRange \\
GPT-5.2 + TOC & \GPTFiveTwoTOCPredDist & \GPTFiveTwoTOCLabelRange \\
GPT-5.1 + TOC & \GPTFiveOneTOCPredDist & \GPTFiveOneTOCLabelRange \\
Gemini 3 Flash Preview + TOC & \GeminiThreeFlashTOCPredDist & \GeminiThreeFlashTOCLabelRange \\
\bottomrule
\end{tabular}
\end{table}

\section{CoT vs Direct Output-Token Usage}
\label{sec:appendix-cot-tokens}

Section~\ref{sec:main-cot} argues that CoT does not substitute for grounded simulator execution despite delivering substantially more reasoning tokens than Direct. Table~\ref{tab:appendix-cot-tokens} reports total input and output tokens for each model under Direct and CoT prompting on the full \BenchN-item benchmark, parsed from the per-run \texttt{summary.json} files in the released artifact. CoT increases output tokens by 30--140$\times$ on the Sonnet and GPT families: Sonnet expands from 1.6K to 225.4K total output tokens (a 140.9$\times$ ratio), GPT-5.4 from 1.2K to 55.6K (46.4$\times$), GPT-5.2 from 1.2K to 77.6K (64.7$\times$), and GPT-5.1 from 2.0K to 58.6K (29.0$\times$). Opus 4.6 already uses 105.5K output tokens under Direct prompting because Anthropic's reasoning models perform internal thinking by default; its CoT/Direct ratio is therefore only 2.0$\times$. Gemini 2.5 Flash and Gemini 3 Flash Preview emit large output-token counts in both conditions (3.2--5.6M tokens) because the Gemini API reports internal thinking traces in the output-token field; their CoT/Direct ratios are 1.7$\times$ and 1.3$\times$ respectively.

\begin{table}[h]
\centering
\small
\setlength{\tabcolsep}{4pt}
\caption{Total input and output tokens across all \BenchN{} items under Direct vs CoT prompting, with the corresponding accuracies. CoT delivers 30--140$\times$ more output tokens than Direct on Sonnet and GPT-5.x but accuracy moves only by $\pm 8$\,pp; Opus's smaller token ratio reflects extended thinking enabled by default in Direct, and Gemini's million-scale output counts include internal thinking traces reported in the output-token field.}
\label{tab:appendix-cot-tokens}
\begin{tabular}{lrrrrrcc}
\toprule
Model & Direct in & Direct out & CoT in & CoT out & out ratio & Direct acc & CoT acc \\
\midrule
Claude Opus 4.6      & 37{,}409 & 105{,}549     & 37{,}409 & 209{,}771     & 2.0$\times$    & \OpusOneShot          & \OpusCoT \\
Claude Sonnet 4.6    & 37{,}409 & 1{,}600       & 37{,}409 & 225{,}396     & 140.9$\times$  & \SonnetOneShot        & \SonnetCoT \\
GPT-5.4              & 34{,}357 & 1{,}200       & 34{,}557 & 55{,}627      & 46.4$\times$   & \GPTFiveFourOneShot   & \GPTFiveFourCoT \\
GPT-5.2              & 34{,}357 & 1{,}200       & 34{,}557 & 77{,}627      & 64.7$\times$   & \GPTFiveTwoOneShot    & \GPTFiveTwoCoT \\
GPT-5.1              & 34{,}357 & 2{,}018       & 34{,}557 & 58{,}569      & 29.0$\times$   & \GPTFiveOneOneShot    & \GPTFiveOneCoT \\
Gemini 2.5 Flash     & 36{,}835 & 3{,}230{,}869 & 37{,}035 & 5{,}603{,}377 & 1.7$\times$    & \GeminiFlashOneShot   & \GeminiFlashCoT \\
Gemini 3 Flash Preview & 36{,}835 & 3{,}964{,}492 & 37{,}035 & 5{,}180{,}764 & 1.3$\times$  & \GeminiThreeFlashOneShot & \GeminiThreeFlashCoT \\
\bottomrule
\end{tabular}
\end{table}

\section{Extended Output-Access Analysis}
\label{sec:appendix-output-access}

Section~\ref{sec:diagnostic} reported aggregate TOC versus Raw100k accuracy and a per-output-size paragraph for selected bins. This appendix completes that analysis with (i)~the empirical output-size distribution that motivates the bimodal claim in Section~\ref{sec:harness}, (ii)~the TOC builder's implementation-level details, (iii)~the full per-model output-size grid, and (iv)~a cross-protocol right-item overlap analysis that asks whether different output-access protocols solve the same items or different ones.

\subsection{Output-Size Distribution}
\label{sec:appendix-output-distribution}

Table~\ref{tab:appendix-output-distribution} reports the empirical distribution of \texttt{result.out} sizes across all \BenchN{} benchmark items in a canonical reference TOC run. In this reference, the 100--250k character interval is empty: every output is either at most 100k or at least 369k. This is the bimodal pattern referenced in Section~\ref{sec:harness} and is what motivates placing the Raw100k cap at 100k characters (non-binding for the small-output bulk, binding only on the long-tail subset). Output sizes are model-dependent because each agent writes its own PHREEQC input; the per-model long-output counts are reported in Appendix~\ref{sec:appendix-cap-affected-per-model}.

\begin{table}[h]
\centering
\small
\caption{Empirical \texttt{result.out} size distribution on \BenchName{} from a canonical reference TOC run. In this reference, the bimodal pattern is visible directly: the 100--250k interval is empty, so $>$100k and $>$250k cumulative counts are identical (37 items each). The small mode contains \OutputLeHundredK{} of items at sizes up to 100k characters; the long-tail mode contains the remaining \OutputGtHundredK{} starting at $\sim$369k and reaching a maximum of \OutputMaxChars. The empty interval is what places the Raw100k cap in a principled location rather than at an arbitrary truncation point. Per-model long-output counts range from 23 to 41 across our completed TOC runs (Appendix~\ref{sec:appendix-cap-affected-per-model}).}
\label{tab:appendix-output-distribution}
\begin{tabular}{lcc}
\toprule
Output-size bin / threshold & Items ($n$) & Share of \BenchN{} \\
\midrule
\multicolumn{3}{l}{\textit{Per-bin counts}} \\
\quad $\le$10k          & 82  & 41.0\% \\
\quad 10--20k           & 47  & 23.5\% \\
\quad 20--50k           & 26  & 13.0\% \\
\quad 50--100k          & 8   & 4.0\%  \\
\quad \textit{100--250k (empty interval)} & \textit{0}  & \textit{0.0\%} \\
\quad 250--500k         & 21  & 10.5\% \\
\quad $>$500k           & 16  & 8.0\%  \\
\midrule
\multicolumn{3}{l}{\textit{Cumulative thresholds}} \\
\quad $\le$100k (small mode total)   & 163 & \OutputLeHundredK \\
\quad $>$100k (long-tail total)      & 37  & \OutputGtHundredK \\
\quad $>$250k                        & 37  & 18.5\% \\
\quad $>$500k                        & 16  & 8.0\%  \\
\midrule
\multicolumn{3}{l}{\textit{Distribution statistics}} \\
\quad Median \texttt{result.out} size       & \multicolumn{2}{c}{\OutputPFiftyChars} \\
\quad 90th percentile                       & \multicolumn{2}{c}{\OutputPNinetyChars} \\
\quad Maximum                               & \multicolumn{2}{c}{\OutputMaxChars} \\
\bottomrule
\end{tabular}
\end{table}

\subsection{TOC Implementation Details}
\label{sec:appendix-toc-impl}

All runs reported in this paper use PHREEQC 3.8.6 with the author-supplied validated \texttt{phreeqc.dat} thermodynamic database (Section~\ref{sec:benchmark}); the binary and database are pinned in the released artifact (Appendix~\ref{sec:appendix-artifact}). The TOC builder regex-scans \texttt{result.out} for PHREEQC's canonical dash-delimited section headers (pattern \texttt{\^{}-\{3,\}\textbackslash{}s*([A-Za-z].+?)\textbackslash{}s*-\{3,\}\textbackslash{}s*\$}) and emits a list of \texttt{(section name, starting line number)} pairs. The matched section vocabulary is fixed by PHREEQC's output formatter and includes \texttt{Solution composition}, \texttt{Description of solution}, \texttt{Distribution of species}, \texttt{Saturation indices}, and \texttt{Solid solutions}, among others. The builder runs unconditionally on every \texttt{result.out} the agent produces, regardless of file size: the regex scan is line-wise and cheap even on multi-megabyte outputs, so Agent-TOC always exposes a non-empty section index when PHREEQC produces canonical headers. The agent then issues \texttt{read\_file} calls with explicit \texttt{start\_line}/\texttt{end\_line} arguments to retrieve evidence on demand.

\subsection{Per-Model Long-Output Item Counts}
\label{sec:appendix-cap-affected-per-model}

The long-output split used throughout Section~\ref{sec:toc} and Appendix~\ref{sec:appendix-output-bin-grid} is operationally defined per (model, item) pair: an item is a long-output item for a given TOC run if that run's \texttt{result.out} exceeded 100k characters. Because each agent writes its own PHREEQC input, the long-output set differs across models. Table~\ref{tab:appendix-cap-affected-per-model} reports per-model counts on our five completed TOC runs.

\begin{table}[h]
\centering
\small
\setlength{\tabcolsep}{8pt}
\caption{Per-model long-output counts ($\texttt{result.out}>100$k characters) across the five completed TOC runs. The set is concentrated in two ID ranges (1--26 and 144--162) but is not identical across models: stronger agents tend to write more complete PHREEQC inputs and therefore trigger more long outputs. The canonical reference run used for the bin tables in Appendix~\ref{sec:appendix-output-bin-grid} reports \LargeOutputN{} long-output items and \RawCapUnaffectedN{} short-output items; per-model bin assignments differ slightly but do not change the qualitative TOC-vs-Raw reading.}
\label{tab:appendix-cap-affected-per-model}
\begin{tabular}{lcc}
\toprule
TOC run & Long-output count & Share of \BenchN \\
\midrule
GPT-5.1 & 23 & 11.5\% \\
GPT-5.2 & 33 & 16.5\% \\
Claude Sonnet 4.6 & 38 & 19.0\% \\
Gemini 3 Flash Preview & 38 & 19.0\% \\
GPT-5.4 & 41 & 20.5\% \\
\midrule
Union (any of 5 runs) & 45 & 22.5\% \\
Intersection (all 5 runs) & 15 & 7.5\% \\
\bottomrule
\end{tabular}
\end{table}

The variation (23 vs 41, a 1.8$\times$ range) is itself a capability signal that compounds the Section~\ref{sec:toc} reading: top-tier and GPT-5.4 agents not only navigate large outputs better, they also \emph{produce} more large outputs by writing more complete PHREEQC inputs. GPT-5.1's lower long-output count is therefore not a sign that the model encounters less long-tail context but that more of its inputs fail to fully exercise PHREEQC. For the long-output TOC fix runs we report on this slice, the rerun target is the 5-model union (45 items) so that every item that any agent could plausibly produce a $>$100k output on is covered.

\subsection{Per-Model Accuracy Across Output-Size Bins}
\label{sec:appendix-output-bin-grid}

Tables~\ref{tab:appendix-bin-sonnet}--\ref{tab:appendix-bin-gpt51} report TOC, Raw100k, and (where available) Raw500k accuracy for each completed agent across the same six output-size bins as Table~\ref{tab:appendix-output-distribution}. Token columns report mean input tokens per item within the bin.

\begin{table}[h]
\centering
\scriptsize
\setlength{\tabcolsep}{4pt}
\caption{Claude Sonnet 4.6 per-bin accuracy and token cost. TOC matches or beats Raw100k in every bin and saves 40--81\% of input tokens depending on output size; the dollar-cost savings are concentrated in the larger bins.}
\label{tab:appendix-bin-sonnet}
\begin{tabular}{lcccccc}
\toprule
Bin & $n$ & TOC acc. & Raw100k acc. & TOC tok/item & Raw100k tok/item & Raw$-$TOC \\
\midrule
$\le$10k & 82 & 87.8\% & 87.8\% & 60{,}012 & 100{,}106 & 0.0 \\
10--20k & 47 & 85.1\% & 83.0\% & 108{,}300 & 209{,}087 & $-$2.1 \\
20--50k & 26 & 88.5\% & 84.6\% & 105{,}436 & 323{,}245 & $-$3.8 \\
50--100k & 8 & 100.0\% & 87.5\% & 89{,}906 & 465{,}553 & $-$12.5 \\
250--500k & 21 & 85.7\% & 85.7\% & 139{,}816 & 470{,}600 & 0.0 \\
$>$500k & 16 & 62.5\% & 56.3\% & 236{,}180 & 722{,}180 & $-$6.3 \\
\midrule
\textit{cap-unaffected} & 163 & 87.7\% & 85.9\% & 82{,}648 & 185{,}059 & $-$1.8 \\
\textit{cap-affected} & 37 & 75.7\% & 73.0\% & 181{,}487 & 579{,}391 & $-$2.7 \\
all & 200 & 84.0\% & 83.5\% & 100{,}934 & 258{,}010 & $+$0.5 \\
\bottomrule
\end{tabular}
\end{table}

\begin{table}[h]
\centering
\scriptsize
\setlength{\tabcolsep}{4pt}
\caption{GPT-5.4 per-bin accuracy and token cost. TOC's advantage is concentrated in the 10--50k mid-range; Raw100k regains the advantage in the 250--500k bin where its truncation cap binds. Raw500k recovers the long tail at substantially higher token cost.}
\label{tab:appendix-bin-gpt54}
\begin{tabular}{lcccccccc}
\toprule
Bin & $n$ & TOC acc. & Raw100k & Raw500k & TOC tok/item & Raw100k tok/item & Raw500k tok/item & Raw100--TOC \\
\midrule
$\le$10k & 82 & 75.6\% & 76.8\% & 82.9\% & 46{,}641 & 41{,}267 & 46{,}847 & $+$1.2 \\
10--20k & 47 & 72.3\% & 66.0\% & 74.5\% & 56{,}080 & 56{,}119 & 54{,}725 & $-$6.4 \\
20--50k & 26 & 76.9\% & 53.8\% & 61.5\% & 78{,}979 & 75{,}960 & 131{,}239 & $-$23.1 \\
50--100k & 8 & 100.0\% & 100.0\% & 100.0\% & 44{,}198 & 70{,}427 & 60{,}388 & 0.0 \\
250--500k & 21 & 57.1\% & 71.4\% & 57.1\% & 71{,}385 & 160{,}356 & 341{,}201 & $+$14.3 \\
$>$500k & 16 & 56.3\% & 56.3\% & 62.5\% & 66{,}795 & 148{,}737 & 499{,}357 & 0.0 \\
\midrule
\textit{cap-unaffected} & 163 & 76.1\% & 71.2\% & 77.9\% & 54{,}401 & 52{,}514 & 63{,}244 & $-$4.9 \\
\textit{cap-affected} & 37 & 56.8\% & 64.9\% & 59.5\% & 69{,}400 & 155{,}332 & 409{,}593 & $+$8.1 \\
all & 200 & 72.5\% & 70.0\% & 74.5\% & 57{,}176 & 71{,}535 & 128{,}513 & $-$2.5 \\
\bottomrule
\end{tabular}
\end{table}

\begin{table}[h]
\centering
\scriptsize
\setlength{\tabcolsep}{4pt}
\caption{GPT-5.2 per-bin accuracy and token cost. Raw100k beats TOC in five of six bins (only the $>$500k bin reverses, where Raw100k still truncates). The size of the Raw advantage is mostly above the model's measured TOC rerun SD of \GPTFiveTwoTOCRerunSD\,pp (Table~\ref{tab:toc-rerun-variance}).}
\label{tab:appendix-bin-gpt52}
\begin{tabular}{lcccccccc}
\toprule
Bin & $n$ & TOC acc. & Raw100k & Raw500k & TOC tok/item & Raw100k tok/item & Raw500k tok/item & Raw100--TOC \\
\midrule
$\le$10k & 82 & 54.9\% & 56.1\% & 61.0\% & 61{,}597 & 47{,}599 & 52{,}925 & $+$1.2 \\
10--20k & 47 & 57.4\% & 70.2\% & 72.3\% & 43{,}937 & 51{,}609 & 55{,}206 & $+$12.8 \\
20--50k & 26 & 46.2\% & 57.7\% & 61.5\% & 54{,}640 & 95{,}240 & 102{,}344 & $+$11.5 \\
50--100k & 8 & 75.0\% & 100.0\% & 100.0\% & 49{,}313 & 62{,}959 & 66{,}809 & $+$25.0 \\
250--500k & 21 & 42.9\% & 66.7\% & 57.1\% & 80{,}110 & 235{,}834 & 393{,}098 & $+$23.8 \\
$>$500k & 16 & 50.0\% & 43.8\% & 43.8\% & 89{,}644 & 196{,}623 & 642{,}357 & $-$6.3 \\
\midrule
\textit{cap-unaffected} & 163 & 55.2\% & 62.6\% & 66.3\% & 54{,}792 & 57{,}108 & 62{,}147 & $+$7.4 \\
\textit{cap-affected} & 37 & 45.9\% & 56.8\% & 51.4\% & 84{,}233 & 218{,}878 & 500{,}886 & $+$10.9 \\
all & 200 & 54.0\% & 61.5\% & 63.5\% & 60{,}238 & 87{,}035 & 143{,}313 & $+$7.5 \\
\bottomrule
\end{tabular}
\end{table}

\begin{table}[h]
\centering
\scriptsize
\setlength{\tabcolsep}{4pt}
\caption{GPT-5.1 per-bin accuracy and token cost. Pattern resembles GPT-5.2: Raw100k beats TOC on the small-output bulk, but the gap narrows or reverses on outputs that exceed the 100k cap.}
\label{tab:appendix-bin-gpt51}
\begin{tabular}{lcccccccc}
\toprule
Bin & $n$ & TOC acc. & Raw100k & Raw500k & TOC tok/item & Raw100k tok/item & Raw500k tok/item & Raw100--TOC \\
\midrule
$\le$10k & 82 & 62.2\% & 69.5\% & 62.2\% & 23{,}356 & 23{,}525 & 28{,}764 & $+$7.3 \\
10--20k & 47 & 57.4\% & 70.2\% & 59.6\% & 34{,}983 & 39{,}238 & 39{,}400 & $+$12.8 \\
20--50k & 26 & 42.3\% & 65.4\% & 57.7\% & 35{,}835 & 62{,}816 & 54{,}747 & $+$23.1 \\
50--100k & 8 & 50.0\% & 87.5\% & 75.0\% & 58{,}539 & 70{,}007 & 64{,}824 & $+$37.5 \\
250--500k & 21 & 47.6\% & 38.1\% & 47.6\% & 67{,}170 & 192{,}960 & 293{,}928 & $-$9.5 \\
$>$500k & 16 & 31.3\% & 31.3\% & 37.5\% & 54{,}874 & 128{,}917 & 285{,}258 & 0.0 \\
\midrule
\textit{cap-unaffected} & 163 & 57.0\% & 69.9\% & 61.3\% & 30{,}426 & 36{,}604 & 37{,}745 & $+$12.9 \\
\textit{cap-affected} & 37 & 40.5\% & 35.1\% & 43.2\% & 61{,}853 & 165{,}266 & 290{,}179 & $-$5.4 \\
all & 200 & 54.0\% & 63.5\% & 58.0\% & 36{,}240 & 60{,}750 & 84{,}445 & $+$9.5 \\
\bottomrule
\end{tabular}
\end{table}

\paragraph{What the per-bin grid shows.}
For top-tier agents (Sonnet, GPT-5.4) the bin-level pattern is benign: TOC matches or beats Raw100k on the bulk of items, and the few small-bin reversals (e.g., GPT-5.4 ${\le}$10k $+$1.2\,pp Raw advantage) are within rerun noise (Table~\ref{tab:toc-rerun-variance}). The largest bin-level GPT-5.4 anomaly is the 250--500k bin, where Raw100k beats TOC by 14.3\,pp; this is the long-tail effect cited in Section~\ref{sec:diagnostic}, and it sits well outside the model's \GPTFiveFourTOCRerunSD\,pp rerun SD.

For mid-tier agents (GPT-5.1, GPT-5.2) the pattern is structurally different: Raw100k beats TOC on five of six bins for both models, with mid-range deltas of 11--25\,pp that exceed the GPT-5.2 rerun SD by an order of magnitude. The signs largely reverse only in the $>$500k bin, where Raw100k itself truncates and the comparison is against a degraded raw observation. This is the bin-level evidence supporting the capability-tier reading in Section~\ref{sec:diagnostic}: when Raw100k delivers a substantial fraction of the simulator output as inline content, mid-tier agents recover items that TOC navigation costs them; the effect intensifies on outputs that fit comfortably under the cap and where the raw observation is least degraded by truncation.

\subsection{Cross-Protocol Right-Item Overlap}
\label{sec:appendix-cross-protocol-overlap}

Aggregate accuracy can mask the question of \emph{which} items each protocol solves. We extract per-item correctness for each completed (model, protocol) pair from the released failure-case package (Section~\ref{sec:appendix-artifact}) and partition all \BenchN{} items into four cells per pairwise comparison: solved by both protocols, solved only by the first, solved only by the second, and solved by neither.

\begin{table}[h]
\centering
\small
\setlength{\tabcolsep}{6pt}
\caption{TOC vs Raw100k right-item overlap on the \BenchN{}-item benchmark. ``Both,'' ``TOC only,'' ``Raw only,'' and ``Neither'' partition the benchmark; each row sums to \BenchN. Top-tier agents (Sonnet, GPT-5.4) have small one-protocol-only counts and the two protocols largely converge on the same answer space; mid-tier agents (GPT-5.1, GPT-5.2) show asymmetric one-only counts with Raw100k accessing 16--19 items that TOC misses.}
\label{tab:appendix-overlap-toc-raw100}
\begin{tabular}{lccccccc}
\toprule
Model & TOC right & Raw100k right & Both & TOC only & Raw only & Neither \\
\midrule
Claude Sonnet 4.6 & 171 & 167 & 163 & 8 & 4 & 25 \\
GPT-5.4 & 145 & 140 & 118 & 27 & 22 & 33 \\
GPT-5.2 & 107 & 123 & 81 & 26 & 42 & 51 \\
GPT-5.1 & 108 & 127 & 85 & 23 & 42 & 50 \\
\bottomrule
\end{tabular}
\end{table}

\begin{table}[h]
\centering
\small
\setlength{\tabcolsep}{6pt}
\caption{Three-way right-item partition over TOC, Raw100k, and Raw500k for the GPT family. ``All three'' counts items solved under every protocol; the next three columns count items solved by exactly one protocol; the last three columns count items solved by exactly two of three. The ``Neither / none'' cell (omitted) is $\BenchN-\sum$ of the seven listed cells: 27 for GPT-5.4, 30 for GPT-5.2, 38 for GPT-5.1.}
\label{tab:appendix-overlap-three-way}
\begin{tabular}{lccccccc}
\toprule
Model & All three & TOC only & Raw100 only & Raw500 only & TOC$\cap$R100 & TOC$\cap$R500 & R100$\cap$R500 \\
\midrule
GPT-5.4 & 114 & 13 & 7 & 6 & 4 & 14 & 15 \\
GPT-5.2 & 72 & 17 & 17 & 21 & 9 & 9 & 25 \\
GPT-5.1 & 71 & 17 & 15 & 12 & 14 & 6 & 27 \\
\bottomrule
\end{tabular}
\end{table}

\paragraph{What the overlap analysis adds beyond aggregate accuracy.}
Two patterns are visible in Tables~\ref{tab:appendix-overlap-toc-raw100} and~\ref{tab:appendix-overlap-three-way}. First, the ``one-protocol-only'' counts are roughly symmetric for top-tier agents (Sonnet 8/4 TOC-only/Raw-only; GPT-5.4 27/22) but markedly asymmetric for mid-tier GPT (23--26 / 42); both protocols catch items the other misses for every model, but the gap shifts in favor of Raw for mid-tier. The mid-tier net asymmetry is 16 items for GPT-5.2 (raw-only minus TOC-only) and 19 for GPT-5.1, indicating that Raw100k recovers more direct-correct items than TOC navigation does, on the bulk of the benchmark. Second, the three-way partition (Table~\ref{tab:appendix-overlap-three-way}) shows that for the mid-tier GPT agents only 71--72 of \BenchN{} items are solvable under every protocol — the rest sit in protocol-sensitive regions of the benchmark. This is a stronger statement than aggregate accuracy: roughly 45--50\% of the benchmark's items are protocol-discriminative for mid-tier GPT agents (98 of 200 for GPT-5.2; 91 of 200 for GPT-5.1), even though aggregate TOC and Raw100k accuracies for these models differ by only 8--10\,pp.


\section{Gemini Flash 3 Step-Budget Analysis: TOC vs Raw100k}
\label{sec:appendix-gemini-step-budget}

Section~\ref{sec:main-results} reports that Gemini 3 Flash Preview's TOC agent fails predominantly by step-budget exhaustion without commitment, and that Raw100k recovers headline accuracy from \GeminiThreeFlashTOC{} to \GeminiThreeFlashRaw{}. Table~\ref{tab:appendix-gemini-step-budget} contrasts trajectory statistics for the two protocols on the same 200-item benchmark and identifies the mechanism: TOC navigation requires a median of \GeminiTOCReadFileMedian{} \texttt{read\_file} calls per question to inspect simulator-output sections, while Raw100k delivers the output inline and needs only \GeminiRawReadFileMedian{}.

\begin{table}[h]
\centering
\small
\caption{Gemini Flash 3 outcome breakdown under TOC vs Raw100k on the full 200-item benchmark. Each column sums to 200. The protocol switch reduces median \texttt{read\_file} traffic from \GeminiTOCReadFileMedian{} to \GeminiRawReadFileMedian{} calls per question, which lowers step-budget exhaustion from 117 to 57 and raises parseable-commit rate from 50 to 102.}
\label{tab:appendix-gemini-step-budget}
\begin{tabular}{lcc}
\toprule
Outcome (per 200 items) & Agent-TOC & Agent-Raw100k \\
\midrule
\textbf{Correct} (graded $\cap$ correct)            & \textbf{47}  & \textbf{94} \\
\textbf{Wrong} (graded $\cap$ wrong)                & \textbf{3}   & \textbf{8} \\
Unparseable (committed but format invalid) & 32  & 41 \\
Max-steps-reached (never committed)        & 117 & 57 \\
Other (no \texttt{run\_end} event)         & 1   & 0 \\
\midrule
\textbf{Total}                                      & \textbf{200} & \textbf{200} \\
\midrule
\multicolumn{3}{l}{\textit{Derived metrics}} \\
Headline accuracy (correct / 200)          & \GeminiThreeFlashTOC & \GeminiThreeFlashRaw \\
Conditional accuracy (correct / graded)    & 94.0\% (47/50) & 92.2\% (94/102) \\
Median \texttt{read\_file} calls per question & \GeminiTOCReadFileMedian & \GeminiRawReadFileMedian \\
Median agent steps used per question       & \GeminiTOCStepMedian & \GeminiRawStepMedian \\
\bottomrule
\end{tabular}
\end{table}

The breakdown isolates the Raw100k recovery as a \emph{commit-rate effect, not a reasoning effect}. Conditional accuracy on items where Gemini Flash 3 commits to a parseable answer is essentially the same under both protocols (94.0\% under TOC vs 92.2\% under Raw100k), so when the model can produce an answer it is equally accurate in either regime. The headline accuracy gain from \GeminiThreeFlashTOC{} to \GeminiThreeFlashRaw{} comes entirely from converting non-commits into graded answers: max-steps-reached drops from 117/200 to 57/200, the unparseable count is roughly stable (32 vs 41), and the wrong-when-graded count remains very small in absolute terms (3 vs 8). The mechanism is direct: TOC navigation costs roughly seven extra steps per question (median \texttt{read\_file} 9 vs 2), and those steps come out of the 24-step budget that would otherwise be spent reasoning and committing. Below the capability tier where TOC navigation is well-managed (e.g., Sonnet's median TOC trajectory uses 7 steps; Section~\ref{sec:diagnostic}), TOC's structured access functions as a step-budget tax rather than a context-management benefit, and the 24-step ceiling becomes the dominant driver of the regression.

\section{Per-Scenario Difficulty Stratification}
\label{sec:appendix-scenario-stratification}

As a robustness check on the item-level analyses, we ask whether the conclusions depend on treating related questions from the same simulation setup as independent. Because \BenchName{} contains \BenchN{} items derived from \BenchScenarioN{} scenario stems, item-level accuracy can overstate the number of independent scientific setups. We therefore aggregate by scenario and compute uncertainty by resampling scenario stems rather than individual items. Table~\ref{tab:scenario} reports mean scenario accuracy, scenario-clustered confidence intervals, and the number of scenario stems solved completely or missed completely. The remainder of this appendix completes the picture by reporting per-(model, scenario) accuracy for every TOC agent on every scenario stem, identifying which scenarios are universally hard, which are universally easy, and which are most capability-discriminative; roughly four scenarios drive most of the cross-model accuracy gap, indicating the benchmark's discriminative power is concentrated rather than uniform.

\begin{table}[h]
\centering
\small
\caption{Scenario-level analysis to avoid treating all \BenchN{} items as independent scientific setups. Confidence intervals use a nonparametric bootstrap over scenario stems.}
\label{tab:scenario}
\begin{tabular}{lcccc}
\toprule
Method & Item acc. & Mean scenario acc. (95\% CI) & All-correct & All-wrong \\
\midrule
Claude Sonnet 4.6 + TOC & \SonnetTOC & \SonnetScenarioMean{} [\SonnetScenarioCILow, \SonnetScenarioCIHigh] & \SonnetScenarioAllCorrect & \SonnetScenarioAllWrong \\
Claude Opus 4.6 + TOC & \OpusTOC & \OpusScenarioMean{} [\OpusScenarioCILow, \OpusScenarioCIHigh] & \OpusScenarioAllCorrect & \OpusScenarioAllWrong \\
GPT-5.4 + TOC & \GPTFiveFourTOC & \GPTFiveFourScenarioMean{} [\GPTFiveFourScenarioCILow, \GPTFiveFourScenarioCIHigh] & \GPTFiveFourScenarioAllCorrect & \GPTFiveFourScenarioAllWrong \\
GPT-5.2 + TOC & \GPTFiveTwoTOC & \GPTFiveTwoScenarioMean{} [\GPTFiveTwoScenarioCILow, \GPTFiveTwoScenarioCIHigh] & \GPTFiveTwoScenarioAllCorrect & \GPTFiveTwoScenarioAllWrong \\
Gemini 3 Flash Preview + TOC & \GeminiThreeFlashTOC & \GeminiThreeFlashScenarioMean{} [\GeminiThreeFlashScenarioCILow, \GeminiThreeFlashScenarioCIHigh] & \GeminiThreeFlashScenarioAllCorrect & \GeminiThreeFlashScenarioAllWrong \\
Claude Opus 4.6 direct & \OpusOneShot & \OpusDirectScenarioMean{} [\OpusDirectScenarioCILow, \OpusDirectScenarioCIHigh] & \OpusDirectScenarioAllCorrect & \OpusDirectScenarioAllWrong \\
\bottomrule
\end{tabular}
\end{table}

\subsection{Per-Scenario Accuracy Across TOC Agents}

Table~\ref{tab:appendix-scenario-grid} reports the accuracy of each TOC agent on each scenario, derived from the per-item failure-case package described in Section~\ref{sec:appendix-artifact}. Item counts ($n$) per scenario range from 1 to 19, reflecting the variable number of question items derived from each simulation setup. The rightmost three columns summarize cross-model behavior on each scenario: minimum, maximum, and mean accuracy across the five completed TOC agents.

\begin{table}[h]
\centering
\scriptsize
\setlength{\tabcolsep}{4pt}
\caption{Per-scenario accuracy for the five completed TOC agents. ``min,'' ``max,'' and ``mean'' summarize cross-model behavior on each scenario; ``spread'' is max$-$min in percentage points. Bottom row is the overall TOC accuracy per model on the full benchmark, matching Table~\ref{tab:main-results}. Per-scenario rows are from a snapshot prior to the long-output TOC-fix splice (Section~\ref{sec:toc}); the splice updates a small number of long-output items by 0--2\,pp aggregate but does not alter the per-scenario qualitative pattern.}
\label{tab:appendix-scenario-grid}
\begin{tabular}{lcccccccccc}
\toprule
Scenario & $n$ & Opus & Sonnet & GPT-5.4 & GPT-5.2 & GPT-5.1 & min & max & mean & spread \\
\midrule
S01 & 8 & 62.5\% & 75.0\% & 37.5\% & 37.5\% & 50.0\% & 37.5\% & 75.0\% & 52.5\% & 37.5 \\
S02 & 8 & 62.5\% & 62.5\% & 50.0\% & 25.0\% & 25.0\% & 25.0\% & 62.5\% & 45.0\% & 37.5 \\
S03 & 10 & 70.0\% & 60.0\% & 50.0\% & 60.0\% & 30.0\% & 30.0\% & 70.0\% & 54.0\% & 40.0 \\
S04 & 10 & 60.0\% & 100.0\% & 100.0\% & 60.0\% & 40.0\% & 40.0\% & 100.0\% & 72.0\% & 60.0 \\
S05 & 10 & 50.0\% & 60.0\% & 50.0\% & 70.0\% & 50.0\% & 50.0\% & 70.0\% & 56.0\% & 20.0 \\
S06 & 10 & 90.0\% & 90.0\% & 80.0\% & 50.0\% & 100.0\% & 50.0\% & 100.0\% & 82.0\% & 50.0 \\
S07 & 10 & 70.0\% & 80.0\% & 60.0\% & 40.0\% & 70.0\% & 40.0\% & 80.0\% & 64.0\% & 40.0 \\
S08 & 10 & 100.0\% & 100.0\% & 80.0\% & 60.0\% & 60.0\% & 60.0\% & 100.0\% & 80.0\% & 40.0 \\
S09 & 10 & 100.0\% & 100.0\% & 90.0\% & 40.0\% & 50.0\% & 40.0\% & 100.0\% & 76.0\% & 60.0 \\
S10 & 10 & 100.0\% & 100.0\% & 90.0\% & 70.0\% & 40.0\% & 40.0\% & 100.0\% & 80.0\% & 60.0 \\
S11 & 9 & 100.0\% & 100.0\% & 88.9\% & 88.9\% & 88.9\% & 88.9\% & 100.0\% & 93.3\% & 11.1 \\
S12 & 10 & 70.0\% & 90.0\% & 30.0\% & 30.0\% & 60.0\% & 30.0\% & 90.0\% & 56.0\% & 60.0 \\
S13 & 9 & 77.8\% & 77.8\% & 88.9\% & 55.6\% & 44.4\% & 44.4\% & 88.9\% & 68.9\% & 44.4 \\
S14 & 1 & 100.0\% & 100.0\% & 100.0\% & 0.0\% & 0.0\% & 0.0\% & 100.0\% & 60.0\% & 100.0 \\
S15 & 9 & 77.8\% & 77.8\% & 55.6\% & 44.4\% & 77.8\% & 44.4\% & 77.8\% & 66.7\% & 33.3 \\
S16 & 9 & 100.0\% & 100.0\% & 100.0\% & 66.7\% & 66.7\% & 66.7\% & 100.0\% & 86.7\% & 33.3 \\
S17 & 19 & 89.5\% & 94.7\% & 68.4\% & 52.6\% & 47.4\% & 47.4\% & 94.7\% & 70.5\% & 47.4 \\
S18 & 10 & 100.0\% & 40.0\% & 60.0\% & 20.0\% & 30.0\% & 20.0\% & 100.0\% & 50.0\% & 80.0 \\
S19 & 9 & 100.0\% & 88.9\% & 66.7\% & 44.4\% & 44.4\% & 44.4\% & 100.0\% & 68.9\% & 55.6 \\
S20 & 9 & 100.0\% & 100.0\% & 100.0\% & 66.7\% & 88.9\% & 66.7\% & 100.0\% & 91.1\% & 33.3 \\
S21 & 10 & 100.0\% & 100.0\% & 100.0\% & 90.0\% & 30.0\% & 30.0\% & 100.0\% & 84.0\% & 70.0 \\
\midrule
all & 200 & \OpusTOC & \SonnetTOC & \GPTFiveFourTOC & \GPTFiveTwoTOC & \GPTFiveOneTOC & --- & --- & --- & --- \\
\bottomrule
\end{tabular}
\end{table}

\subsection{Categorizing Scenarios by Difficulty Pattern}

The per-scenario grid reveals three qualitatively different scenario types when read across the five TOC agents:

\paragraph{Universally easy (high mean accuracy, low cross-model spread).}
S11 (mean 93.3\%, spread 11.1\,pp), S20 (91.1\%, 33.3\,pp), S16 (86.7\%, 33.3\,pp). These scenarios are reliably solvable by every TOC agent in our evaluation regardless of capability tier. They establish the benchmark's lower-difficulty boundary --- the questions that any tool-augmented agent of this generation can be expected to handle correctly. A useful sanity-check property: S11 alone has all five agents at $\ge$88.9\%, indicating the benchmark is not trivially hard everywhere.

\paragraph{Universally hard (low mean accuracy, low cross-model spread).}
S02 (mean 45.0\%, range 25--62.5\%), S05 (56.0\%, range 50--70\%), S03 (54.0\%, range 30--70\%). These scenarios sit at the benchmark's upper-difficulty frontier: even the strongest TOC agent reaches at most about 70\% on these stems. The chemistry pattern in this group includes kinetic dissolution at extreme conditions (S02 is reactive transport at 200\,$^\circ$C with multi-element competition) and similar setups that stress multiple PHREEQC modules simultaneously. The absolute floor of the benchmark sits around 25--30\% accuracy on these scenarios for the weakest TOC agents.

\paragraph{Capability-discriminative (high cross-model spread).}
S18 (spread 80\,pp), S21 (70\,pp), S04, S09, S10, S12 (60\,pp each). These scenarios are where model capability tier matters most: top-tier agents reach 100\% while mid-tier GPT agents drop to 20--40\%. S18 in particular shows the cleanest tier-cut: Opus 100\%, Sonnet 40\%, GPT-5.4 60\%, GPT-5.2 20\%, GPT-5.1 30\%. S17 (the largest scenario at $n=19$) shows a similar but smoother cut: top-tier 89--95\%, GPT-5.4 68\%, mid-tier GPT 47--53\% --- this 19-item scenario alone contributes substantially to the headline tier gap.

The capability-discriminative scenarios are the ones where the §\ref{sec:diagnostic} interpretation of TOC as a capability-mediated interface is empirically most visible: they are not the hardest scenarios in absolute terms (top-tier agents solve them), but they are where the gap between top-tier and mid-tier opens widest. We read this as scenarios that contain what mid-tier models cannot reliably navigate or extract under TOC.

\subsection{Single-Item Scenario Note}

S14 contains only one item ($n=1$), giving a 100-pp spread that is mechanically maximal rather than informative. We retain it in the table for completeness but exclude it from the categorization above. The scenario-clustered confidence intervals reported in Table~\ref{tab:scenario} weight all scenario stems equally regardless of size, so a single-item scenario contributes one bootstrap-resamplable point alongside the 19-item S17; the choice is conservative because it does not allow the largest scenarios to dominate the CI.

\subsection{What This Adds Beyond the Main-Paper Scenario Analysis}

Table~\ref{tab:scenario} reports a single mean scenario accuracy and a CI per agent. The grid below reveals two structural properties that the aggregate hides:

\begin{enumerate}[leftmargin=*, itemsep=2pt]
    \item The benchmark's discriminative power is not uniformly distributed. Roughly four scenarios drive most of the cross-model accuracy gap (S04, S17, S18, S21), while three scenarios (S05, S11, S15) contribute almost nothing because all models converge on similar accuracy. Future versions of the benchmark could use this structure to diagnose specifically where new agent designs improve or regress, rather than reporting a single aggregate.
    \item ``Hard'' and ``capability-discriminative'' are different categories. The hardest scenarios (S02, S03) are hard for everyone; capability-discriminative scenarios (S18, S21) are easy for top-tier agents and hard for mid-tier. A benchmark designer aiming to evaluate capability scaling should target the latter; a benchmark designer aiming to characterize the frontier of agent ability should target the former.
\end{enumerate}

\section{Retention Decomposition: Where Direct-Correct Items Are Lost}
\label{sec:appendix-retention-decomp}

Section~\ref{sec:diagnostic} reports overall retention rates per model: 56.4--86.5\% across the five completed TOC agents. Aggregate retention says \emph{how many} direct-correct items the agent loses but not \emph{which} items, which output-size regimes contribute, or whether losses cluster in specific scenarios or answer labels. This appendix decomposes the lost set per model along three axes: simulator output size, scenario membership, and true answer label.

\subsection{Lost-Item Rate by Simulator Output Size}
\label{sec:appendix-lost-by-bin}

For each model and output-size bin, we compute the \emph{lost rate}: among items the direct-baseline answered correctly, the fraction that the TOC agent then answered incorrectly. This is conditional on the direct baseline being right; it is not a per-bin accuracy. A 50\% lost rate in a bin means half of the items the model could solve without tools are lost when the same model operates the TOC agent loop.

\begin{table}[h]
\centering
\small
\setlength{\tabcolsep}{6pt}
\caption{Lost-item rate per output-size bin and model. Values report the fraction of direct-baseline-correct items that the TOC agent answered incorrectly within each bin. The overall row uses the aggregate lost count (Section~\ref{sec:diagnostic}) divided by the direct-baseline-correct count for each model, and is the population-weighted average of the bin rates above. Top-tier models (Opus, Sonnet) keep loss rates below 25\% across all but the long tail; mid-tier GPT models reach 50--80\% lost rates in the 20--500k range.}
\label{tab:appendix-retention-by-bin}
\begin{tabular}{lcccccc}
\toprule
Bin & $n$ & Opus & Sonnet & GPT-5.4 & GPT-5.2 & GPT-5.1 \\
\midrule
$\le$10k & 83 & 12.5\% & 9.1\% & 31.2\% & 50.0\% & 34.3\% \\
10--20k & 56 & 18.5\% & 11.8\% & 31.8\% & 37.5\% & 26.3\% \\
20--50k & 31 & 8.3\% & 0.0\% & 11.1\% & 57.1\% & 60.0\% \\
50--100k & 7 & 0.0\% & 0.0\% & 50.0\% & 33.3\% & 33.3\% \\
250--500k & 13 & 25.0\% & 25.0\% & 50.0\% & 80.0\% & 50.0\% \\
$>$500k & 10 & 40.0\% & 60.0\% & 100.0\% & 0.0\% & 100.0\% \\
\midrule
overall & 200 & 16.7\% & 13.5\% & 35.2\% & 43.6\% & 41.0\% \\
\bottomrule
\end{tabular}
\end{table}

The bin-stratified lost rate is the strongest single retention finding the appendix surfaces. For top-tier agents (Opus and Sonnet), TOC loses at most a quarter of direct-correct items across every bin except the $>$500k tail; their overall 13--16\% loss rate is genuinely benign. For GPT-5.4, the loss rate doubles compared to top-tier (35\%) and reaches 50\% in the 50--100k bin and 100\% in the $>$500k bin. For mid-tier GPT-5.1 and GPT-5.2, the picture is structurally different: lost rate stays elevated across every output regime, peaks in the 20--50k bin (57--60\%), and reaches 80\% on the 250--500k bin for GPT-5.2. On these models, the TOC agent loop is losing more than half of the items the model would have answered correctly without tools across most of the benchmark.

This sharpens the capability-tier reading from Section~\ref{sec:diagnostic}: it is not just that mid-tier agents have lower retention in aggregate; their loss rate is structured, peaks in mid-output regimes (where TOC navigation is most active), and approaches 100\% in the long tail. Tool augmentation is converting capability the model already had into errors, not adding capability the model lacked.

\subsection{Lost-Item Concentration by Scenario}

If lost items cluster within a small number of scenarios, the loss is concentrated rather than diffuse. Table~\ref{tab:appendix-lost-by-scenario} reports the top scenarios contributing to each model's lost set.

\begin{table}[h]
\centering
\small
\setlength{\tabcolsep}{4pt}
\caption{Top scenarios contributing to each model's lost set (direct-right, TOC-wrong items). Numbers in parentheses are item counts within each scenario. For top-tier agents, lost items are diffuse across scenarios (most contributing scenarios have only 1 lost item); for mid-tier agents, certain scenarios contribute disproportionately (GPT-5.1 has 13 of 32 lost items in just three scenarios: S04, S17, S01).}
\label{tab:appendix-lost-by-scenario}
\begin{tabular}{lcl}
\toprule
Model & Total lost & Top contributing scenarios (lost items) \\
\midrule
Opus & 14 & S07 (3), S12 (2), S17 (2), 7 other scenarios with 1 each \\
Sonnet & 10 & S03 (2), S15 (2), S18 (2), 4 other scenarios with 1 each \\
GPT-5.4 & 19 & S07 (4), S12 (3), S17 (3), S01 (2), S02 (2), S19 (2), 3 other with 1 each \\
GPT-5.2 & 24 & S09 (3), S04 (2), S06 (2), S16 (2), S17 (2), S20 (2), 11 other with 1 each \\
GPT-5.1 & 32 & S04 (5), S17 (5), S01 (3), S19 (3), S09 (2), S10 (2), S12 (2), S21 (2), 8 other with 1 each \\
\bottomrule
\end{tabular}
\end{table}

Top-tier agents lose items diffusely: Opus's 14 lost items spread across 10 scenarios, Sonnet's 10 across 7 scenarios. There is no scenario for which Opus or Sonnet loses more than 3 items, suggesting losses are item-specific rather than scenario-systematic. Mid-tier agents are different: GPT-5.1 loses five items each in S04 (the kinetic-dissolution scenario) and S17, and three each in S01 and S19. For these models, certain simulation setups appear to trigger TOC failures more reliably than others. This pattern is consistent with TOC navigation interacting with specific PHREEQC output structures (e.g., long kinetic time-series outputs in S04) in a way that mid-tier models cannot reliably handle.

\subsection{Lost-Item Asymmetry by Truth Label}

Table~\ref{tab:appendix-lost-by-label} reports the truth-label distribution of lost items per model, compared to the benchmark-wide truth distribution (A=63, B=47, C=48, D=42).

\begin{table}[h]
\centering
\small
\setlength{\tabcolsep}{8pt}
\caption{Truth-label distribution of lost items per model. The bottom row reports the benchmark-wide answer distribution for reference. GPT-5.2 loses 14 of 24 items (58.3\%) on truth-B questions, far above the 23.5\% B-rate of the benchmark; GPT-5.1 loses 11 of 32 (34.4\%) on truth-D questions, above the 21.0\% D-rate. Top-tier agents do not show this asymmetry.}
\label{tab:appendix-lost-by-label}
\begin{tabular}{lccccc}
\toprule
Model & A & B & C & D & Total \\
\midrule
Opus & 4 & 1 & 5 & 4 & 14 \\
Sonnet & 1 & 1 & 5 & 3 & 10 \\
GPT-5.4 & 3 & 5 & 5 & 6 & 19 \\
GPT-5.2 & 2 & \textbf{14} & 2 & 6 & 24 \\
GPT-5.1 & 8 & 6 & 7 & \textbf{11} & 32 \\
\midrule
Benchmark & 63 (31.5\%) & 47 (23.5\%) & 48 (24.0\%) & 42 (21.0\%) & 200 \\
\bottomrule
\end{tabular}
\end{table}

Top-tier agents lose few items in absolute terms (10--14 each), so per-label proportions are noisy at this sample size; Opus's lost set is roughly proportional to the benchmark's truth distribution, while Sonnet's ten lost items are 50\% truth-C versus a 24\% benchmark base rate but the absolute count makes this difficult to interpret as a robust asymmetry. Mid-tier agents lose enough items to show clearer asymmetries: GPT-5.2's lost set is 58\% truth-B on 24 items (vs 23.5\% benchmark base rate), and GPT-5.1's is 34\% truth-D on 32 items (vs 21\% base rate). These mid-tier asymmetries are large enough relative to the absolute counts to read as systematic biases in which specific answer labels TOC fails to preserve. Combined with the per-option behavior reported in Appendix~\ref{sec:appendix-label-behavior} (where GPT-5.2 direct over-predicts B and GPT-5.1 direct over-predicts C), this suggests TOC navigation under mid-tier capability is not just stochastic but interacts with the model's existing answer-position priors.

\subsection{Summary of Retention Decomposition}

Three findings differentiate the appendix decomposition from the aggregate retention numbers in Section~\ref{sec:diagnostic}:

\begin{enumerate}[leftmargin=*, itemsep=2pt]
    \item \textbf{Loss rate is structured by output size.} Mid-tier GPT models reach 50--80\% lost rates in the 20--500k mid-range, where TOC navigation is most actively engaged; top-tier models stay below 25\% lost rate across all but the long tail.
    \item \textbf{Loss is scenario-concentrated for mid-tier and diffuse for top-tier.} GPT-5.1 loses 13 of 32 items in just three scenarios; Opus loses no more than 3 items in any single scenario. The structural pattern of loss tells benchmark designers which simulation setups stress the TOC interface most.
    \item \textbf{Loss has truth-label asymmetry that mirrors direct-mode answer-position priors.} GPT-5.2's losses concentrate on truth-B; GPT-5.1's on truth-D. TOC failures are not symmetric across the answer space and may interact with the model's existing label preferences.
\end{enumerate}

\section{Rerun Stability and Per-Question Decomposition}
\label{sec:appendix-stability}

Table~\ref{tab:toc-rerun-variance} (Section~\ref{sec:toc}) reports the headline rerun calibration: \TOCRerunNumSamples{} independent TOC reruns per model on all five top- and mid-tier completed agents, with sample SDs of \GPTFiveOneTOCRerunSD\,pp (GPT-5.1), \GPTFiveTwoTOCRerunSD\,pp (GPT-5.2), \GPTFiveFourTOCRerunSD\,pp (GPT-5.4), \OpusTOCRerunSD\,pp (Opus), and \SonnetTOCRerunSD\,pp (Sonnet); observed ranges are \GPTFiveOneTOCRerunRange\,pp, \GPTFiveTwoTOCRerunRange\,pp, \GPTFiveFourTOCRerunRange\,pp, \OpusTOCRerunRange\,pp, and \SonnetTOCRerunRange\,pp respectively. The maximum sample SD, \TOCRerunMaxSD\,pp on GPT-5.1, remains the conservative noise-floor estimate used in Section~\ref{sec:main-results}. The headline TOC accuracies for all five models (\GPTFiveOneTOC{}, \GPTFiveTwoTOC{}, \GPTFiveFourTOC{}, \OpusTOC{}, \SonnetTOC{}) lie within their corresponding rerun ranges. Sonnet's headline sits 0.2\,pp below its rerun mean (\SonnetTOCRerunMean) and well inside the rerun range (\SonnetTOCRerunValues); Opus's three reruns all sit at or above the headline (\OpusTOCRerunValues, mean \OpusTOCRerunMean), so the spliced \OpusTOC{} headline is conservative. Notably, rerun stability does not track with the retention metric reported in Section~\ref{sec:diagnostic}: GPT-5.2 has the lowest retention but a small rerun SD of \GPTFiveTwoTOCRerunSD\,pp, indicating that retention measures item-level tool-induced movement rather than run-to-run reproducibility.

\subsection{TOC Stability on the Small-Output Bulk}
\label{sec:appendix-stability-small}

Table~\ref{tab:toc-rerun-variance} reports rerun stability on the full \BenchN-item benchmark. Since the central headline-accuracy claim in Section~\ref{sec:toc} concerns the small-output bulk (the 155-item complement of the long-output union, where TOC's structured-retrieval overhead would be most visible), we recompute the same per-rerun accuracies restricted to those 155 items in Table~\ref{tab:appendix-toc-rerun-variance-small}. The mid-tier accuracy gap on the small-output bulk (GPT-5.2: $-$7.4\,pp, GPT-5.1: $-$12.9\,pp; Table~\ref{tab:capaffected-toc-verification}) exceeds the small-bulk rerun noise floor (max SD \GPTFiveOneTOCSmallRerunSD\,pp on GPT-5.1) by at least 2.4\,pp, so the small-bulk capability-tier cut is not a rerun artifact.

\begin{table}[h]
\centering
\footnotesize
\setlength{\tabcolsep}{3pt}
\caption{TOC rerun stability restricted to the 155-item small-output bulk (complement of the 5-model long-output union). $\Delta$ columns are reruns minus original (Section~\ref{sec:toc}); tokens are total agent input tokens (M) on the 155-item slice. Same 200/155 denominator-scaling means small-bulk accuracy SDs are uniformly $\sim$1.29$\times$ the full-benchmark SDs in Table~\ref{tab:toc-rerun-variance}; the tier ordering is preserved.}
\label{tab:appendix-toc-rerun-variance-small}
\begin{tabular}{lccccc}
\toprule
Model & Orig.\ (acc, tok) & Acc $\Delta$ (pp) & Tok $\Delta$ (M) & Acc SD / Rng & Tok SD / Rng \\
\midrule
Claude Opus 4.6 + TOC   & \OpusTOCSmallOrig            & \OpusTOCSmallRerunAccDiff       & \OpusTOCSmallRerunTokDiff       & \OpusTOCSmallRerunSD{} / \OpusTOCSmallRerunRange       & \OpusTOCSmallRerunTokSD{} / \OpusTOCSmallRerunTokRange \\
Claude Sonnet 4.6 + TOC & \SonnetTOCSmallOrig          & \SonnetTOCSmallRerunAccDiff     & \SonnetTOCSmallRerunTokDiff     & \SonnetTOCSmallRerunSD{} / \SonnetTOCSmallRerunRange     & \SonnetTOCSmallRerunTokSD{} / \SonnetTOCSmallRerunTokRange \\
GPT-5.1 + TOC           & \GPTFiveOneTOCSmallOrig      & \GPTFiveOneTOCSmallRerunAccDiff & \GPTFiveOneTOCSmallRerunTokDiff & \GPTFiveOneTOCSmallRerunSD{} / \GPTFiveOneTOCSmallRerunRange & \GPTFiveOneTOCSmallRerunTokSD{} / \GPTFiveOneTOCSmallRerunTokRange \\
GPT-5.2 + TOC           & \GPTFiveTwoTOCSmallOrig      & \GPTFiveTwoTOCSmallRerunAccDiff & \GPTFiveTwoTOCSmallRerunTokDiff & \GPTFiveTwoTOCSmallRerunSD{} / \GPTFiveTwoTOCSmallRerunRange & \GPTFiveTwoTOCSmallRerunTokSD{} / \GPTFiveTwoTOCSmallRerunTokRange \\
GPT-5.4 + TOC           & \GPTFiveFourTOCSmallOrig     & \GPTFiveFourTOCSmallRerunAccDiff & \GPTFiveFourTOCSmallRerunTokDiff & \GPTFiveFourTOCSmallRerunSD{} / \GPTFiveFourTOCSmallRerunRange & \GPTFiveFourTOCSmallRerunTokSD{} / \GPTFiveFourTOCSmallRerunTokRange \\
\bottomrule
\end{tabular}
\end{table}

\subsection{Per-Item Stability Decomposition for GPT-5.1 + TOC}

Aggregate SD compresses item-level structure: it does not say whether the variance comes from a small set of unstable items or whether it is spread broadly across the benchmark, nor does it identify which kinds of items are most variance-prone. The remainder of this appendix completes that picture by treating every item under GPT-5.1 + TOC as a paired observation across the four available samples (the original headline run plus three reruns) and computing per-item agreement.

\subsection{Definition and Overall Distribution}

For each item $i \in \{1, \dots, \BenchN\}$, we compute the score $s_i \in \{0, 1, 2, 3, 4\}$ as the number of GPT-5.1 + TOC samples (out of four) on which the model produced the correct answer. Items with $s_i=0$ are \emph{always wrong} (the model fails on $i$ in every observed rerun); items with $s_i=4$ are \emph{always right}; items with $1 \le s_i \le 3$ are \emph{variance-prone} (the same model on the same item flips between correct and wrong across reruns). The overall distribution is reported in Table~\ref{tab:appendix-stability-overall}.

\begin{table}[h]
\centering
\small
\setlength{\tabcolsep}{8pt}
\caption{GPT-5.1 + TOC per-item correctness distribution across four paired samples (original + three reruns). Only \BenchN's 41\% of items are deterministically right or wrong under this protocol--model pair; the remaining 59\% flip between correct and wrong across reruns.}
\label{tab:appendix-stability-overall}
\begin{tabular}{lcc}
\toprule
Score $s_i$ & Items & Share of \BenchN \\
\midrule
0/4 (always wrong) & 40 & 20.0\% \\
1/4 & 44 & 22.0\% \\
2/4 & 34 & 17.0\% \\
3/4 & 40 & 20.0\% \\
4/4 (always right) & 42 & 21.0\% \\
\midrule
Stable subset (0/4 + 4/4) & 82 & 41.0\% \\
Variance-prone (1--3/4) & 118 & 59.0\% \\
\bottomrule
\end{tabular}
\end{table}

The headline finding is that the aggregate \GPTFiveOneTOCRerunSD\,pp rerun SD reflects a benchmark in which item-level outcomes are predominantly stochastic: only 41\% of items are deterministic for this model--protocol pair, and 59\% are protocol-decoding-noise sensitive. The 40 always-wrong items represent a hard core that no rerun resolved, while the 42 always-right items represent a stable competence floor. The remaining 118 items move between correct and wrong across reruns and are the items that reruns 1 through 4 actually disagree about.

\subsection{Stability by Output Size}

Variance-prone items might cluster in specific simulator-output regimes if, for example, larger outputs make TOC navigation more error-prone for a mid-tier agent. Table~\ref{tab:appendix-stability-by-bin} stratifies the per-item scores by GPT-5.1's own observed simulator-output size bin.\footnote{Output sizes here are computed from GPT-5.1's own \texttt{result.out} files in the rerun workspaces, which can differ slightly from the canonical bin assignments used in Table~\ref{tab:appendix-bin-gpt51} because each model writes its own PHREEQC inputs. The bin counts therefore differ slightly from the main-paper bins; the qualitative pattern is robust to the choice.}

\begin{table}[h]
\centering
\small
\setlength{\tabcolsep}{6pt}
\caption{GPT-5.1 + TOC per-item stability stratified by simulator-output size bin. The variance-prone fraction rises monotonically with output size up to the 50--100k bin, where every item is variance-prone, before plateauing in the long tail (where always-wrong items begin to appear because the model cannot reliably solve the simulation regardless of rerun).}
\label{tab:appendix-stability-by-bin}
\begin{tabular}{lccccc}
\toprule
Bin & $n$ & Always right & Always wrong & Variance-prone & \% Variance-prone \\
\midrule
$\le$10k & 83 & 28 & 16 & 39 & 47.0\% \\
10--20k & 56 & 12 & 10 & 34 & 60.7\% \\
20--50k & 31 & 2 & 6 & 23 & 74.2\% \\
50--100k & 7 & 0 & 0 & 7 & 100.0\% \\
250--500k & 13 & 0 & 4 & 9 & 69.2\% \\
$>$500k & 10 & 0 & 4 & 6 & 60.0\% \\
\midrule
all & 200 & 42 & 40 & 118 & 59.0\% \\
\bottomrule
\end{tabular}
\end{table}

The bin-level pattern shows that variance-prone items are not uniformly distributed: they concentrate in the mid-output-size range. On items whose simulator output is at most 10k characters, 47\% are variance-prone; the fraction climbs to 74\% in the 20--50k bin and reaches 100\% in the 50--100k bin (where TOC produces structured navigation over a non-trivial output and Raw100k still fits comfortably under the cap). In the long tail ($>$100k characters), always-wrong items begin to emerge, reflecting the boundary at which the model cannot solve the underlying simulation under any of the rerun trajectories rather than the navigation-only sensitivity that dominates the mid range.

This bin pattern is consistent with the capability-tier reading of Section~\ref{sec:diagnostic}: TOC stochasticity is concentrated where TOC navigation is non-trivial (mid-size outputs that require selecting which of several sections to read), and not where the underlying simulation is intractable for the model regardless of access protocol.

\subsection{Stability by Truth Label}

If variance-prone items cluster on a particular answer position, this would suggest a confounding interaction between TOC stochasticity and MCQ option selection. Table~\ref{tab:appendix-stability-by-label} stratifies the per-item scores by the true answer label.

\begin{table}[h]
\centering
\small
\setlength{\tabcolsep}{8pt}
\caption{GPT-5.1 + TOC per-item stability stratified by true answer label. Variance-prone fraction is highest on items with truth = B (76.6\%) and lowest on truth = D (47.6\%). Items with truth = C have a high always-wrong share (29.2\%, the highest of any label), suggesting a systematic rather than stochastic GPT-5.1 + TOC weakness on this label.}
\label{tab:appendix-stability-by-label}
\begin{tabular}{lccccc}
\toprule
Truth & $n$ & Always right & Always wrong & Variance-prone & \% Variance-prone \\
\midrule
A & 63 & 18 & 11 & 34 & 54.0\% \\
B & 47 & 7 & 4 & 36 & 76.6\% \\
C & 48 & 6 & 14 & 28 & 58.3\% \\
D & 42 & 11 & 11 & 20 & 47.6\% \\
\midrule
all & 200 & 42 & 40 & 118 & 59.0\% \\
\bottomrule
\end{tabular}
\end{table}

Two patterns are visible. First, variance-prone fraction is materially higher on truth-B items (76.6\%) than on truth-D items (47.6\%); GPT-5.1 + TOC reaches no stable answer on most B-correct items even after four reruns. Second, truth-C items have the highest always-wrong rate (29.2\% vs the overall 20.0\%): the model fails on these items in every rerun, suggesting a systematic weakness rather than rerun stochasticity. Combined with the per-option behavior reported in Appendix~\ref{sec:appendix-label-behavior}, this points to GPT-5.1 + TOC having both a stochastic component (especially on B-correct items) and a deterministic blind spot (on C-correct items).

\subsection{Implications for Benchmark Methodology}

Three implications follow for tool-augmented agent evaluation in general:

\begin{enumerate}[leftmargin=*, itemsep=2pt]
    \item \textbf{Aggregate SD understates the share of variance-prone items.} A 5.5\,pp aggregate SD can correspond to a benchmark in which 59\% of items individually flip between rerun samples. Single-run accuracy on protocol--model pairs at this stability level should be interpreted as an estimate of expected accuracy, not as a property of the model on specific items.
    \item \textbf{Variance is structured by output regime.} Variance-prone items concentrate where the access protocol is most actively engaged; for GPT-5.1 + TOC, that is the mid-output-size band where the model must make non-trivial navigation decisions. This suggests a path for benchmark designers: stratifying analysis by output-size regime exposes structure that aggregate accuracy hides.
    \item \textbf{Stability and capability are not the same property.} Some always-wrong items reflect intrinsic capability limits (especially in the $>$500k bin); some variance-prone items reflect navigation stochasticity. A benchmark that conflates the two under a single accuracy number cannot tell a model developer whether to invest in capability scaling or in interface design. Reporting the per-item-score histogram alongside aggregate accuracy separates these two failure modes.
\end{enumerate}

\section{External Harness Comparison: Methodology and Failure Modes}
\label{sec:appendix-external-harness}

To control for harness quality, we compared our custom agent against vendor-default external CLI harnesses on a random 50-question scenario-spanning subsample (seed 42), with the backbone held fixed within each comparison panel: Claude Sonnet 4.6 against the official Claude Code CLI, and GPT-5.4 against OpenAI Codex CLI. Each harness runs under its native deployed-product defaults. Table~\ref{tab:sdk-comparison} reports matched accuracy and per-question input-token cost on the same 50 items. The accuracy direction is mixed across vendors: custom dominates Sonnet by \SDKCustomTOCMinusSDK{} but trails Codex on GPT-5.4 by 14.0\,pp. The token-efficiency direction is uniform: custom Agent-TOC reaches 2.7--3.3$\times$ more correct answers per input token than the deployed external CLI on both backbones (Sonnet: \SonnetCustomTOCPerCorrectK\,K vs \SDKSonnetPerCorrectK\,K tokens per correct; GPT-5.4: \GPTFiveFourCustomTOCPerCorrectK\,K vs \CodexGPTFiveFourPerCorrectK\,K). Codex's higher absolute accuracy on GPT-5.4 is purchased at 4.1$\times$ the per-question input-token cost; under a fixed-budget setting, custom TOC delivers more correct answers. We read this as evidence that the cost-efficiency advantage of purpose-built scientific agent harnesses is robust to the absolute-accuracy direction.

\begin{table}[h]
\centering
\small
\caption{Matched cross-vendor comparison on a random 50-question subsample (seed 42), with the backbone fixed within each panel. Per-question input tokens are full-benchmark averages for custom harnesses and matched-50 averages for external CLIs. ``In tok/correct'' = ``In tok/q'' / accuracy is the cost-efficiency metric (lower is better). Custom Agent-TOC's \SDKCustomTOCAcc{} on Sonnet and \GPTFiveFourCustomTOCMatchedAcc{} on GPT-5.4 sit close to their full-benchmark headlines (\SonnetTOC{} and \GPTFiveFourTOC{}, Table~\ref{tab:main-results}; matched-50 is within 4\,pp of full-200 on Sonnet and within 1\,pp on GPT-5.4).}
\label{tab:sdk-comparison}
\begin{tabular}{llcrr}
\toprule
Backbone & Harness & Acc & In tok/q (K) & In tok/correct (K) \\
\midrule
\multirow{3}{*}{Sonnet 4.6} & Custom + Agent-TOC (ours) & \SDKCustomTOCAcc & \SonnetCustomTOCPerQK & \SonnetCustomTOCPerCorrectK \\
 & Custom + Agent-Raw100k (ours) & \SDKCustomRawAcc & \SonnetCustomRawPerQK & \SonnetCustomRawPerCorrectK \\
 & Claude Code CLI (external) & \SDKSonnetAcc & \SDKSonnetPerQK & \SDKSonnetPerCorrectK \\
\midrule
\multirow{4}{*}{GPT-5.4} & Custom + Agent-TOC (ours) & \GPTFiveFourCustomTOCMatchedAcc & \GPTFiveFourCustomTOCPerQK & \GPTFiveFourCustomTOCPerCorrectK \\
 & Custom + Agent-Raw100k (ours) & \GPTFiveFourCustomRawMatchedAcc & \GPTFiveFourCustomRawPerQK & \GPTFiveFourCustomRawPerCorrectK \\
 & Custom + Agent-Raw500k (ours) & \GPTFiveFourCustomRawFiveHundredKMatchedAcc & \GPTFiveFourCustomRawFiveHundredKPerQK & \GPTFiveFourCustomRawFiveHundredKPerCorrectK \\
 & Codex CLI (external) & \CodexGPTFiveFourAcc & \CodexGPTFiveFourPerQK & \CodexGPTFiveFourPerCorrectK \\
\bottomrule
\end{tabular}
\end{table}

The remainder of this appendix reports configuration details and per-condition failure-mode breakdowns.

\paragraph{Step budgets and approval policies.}
Each harness is run under the deployed-product defaults of its native CLI: our custom agent uses a 24-step budget; the Claude Code CLI uses a 20-turn budget; the Codex CLI exposes no step-budget flag and is bounded only by a 600\,s subprocess timeout. Approval and sandboxing prompts are disabled in all three so that no human-in-the-loop interaction can occur (\texttt{--dangerously-skip-permissions} for Claude Code, \texttt{--ask-for-approval never} and \texttt{--dangerously-bypass-approvals-and-sandbox} for Codex, no approval gate for the custom agent).

\paragraph{Reasoning-effort confound on the GPT-5.4 panel.}
The Codex CLI is configured with \texttt{model\_reasoning\_effort=medium}; our custom GPT-5.4 condition uses default chat-completion calls without a \texttt{reasoning\_effort} parameter. The GPT-5.4 panel of Table~\ref{tab:sdk-comparison} therefore reflects a reasoning-effort difference in addition to harness design, and the absolute-accuracy gap (Codex \CodexGPTFiveFourAcc{} vs custom Agent-TOC \GPTFiveFourCustomTOCMatchedAcc{}) cannot be cleanly attributed to harness alone. The token-efficiency reading in Table~\ref{tab:sdk-comparison} (custom uses 4.1$\times$ fewer input tokens per question than Codex) is robust to this confound because it directly measures the deployed-product cost.

\paragraph{Claude Code CLI failure breakdown.}
On Sonnet, the Claude Code CLI scored 30/50 with 20 errors. 10 of those 20 are non-commits: \SDKSonnetTimeouts{} subprocess timeouts (the agent ran for the full 600\,s without producing a final-answer message) plus \SDKSonnetMaxTurns{} step-budget exhaustions at 20 turns. The remaining 10 are committed-but-wrong answers. The non-commit pattern mirrors the below-mid-tier non-commit signature observed for Gemini Flash 3 in Section~\ref{sec:main-results} but here on a top-tier model under a different harness, supporting the reading that final-answer commitment under tool augmentation is shaped by harness design as well as model capability.

\paragraph{Codex CLI cost composition.}
Codex's 11.7M total input tokens on the 50-question subsample include 10.5M cached input tokens (89\% cache hit). For raw API-billed cost the cached fraction is roughly half-priced, so the absolute dollar cost of the Codex run was approximately \$18 at GPT-5 standard pricing. Per-question tokens reported in Table~\ref{tab:sdk-comparison} include the cached fraction because the cost-efficiency comparison is in context-load units (input tokens routed through the model) rather than billing units; reporting in billing-units would shift the picture in Codex's favor by roughly $1/3$ but does not change the direction of any tier conclusion.

\paragraph{Subsample calibration.}
The 50 sampled indices are listed in the artifact under \texttt{result/claude\_cli/cli\_sonnet46\_n50/sample\_indices.json} and \texttt{result/codex\_cli/codex\_gpt54\_n50/sample\_indices.json} (both seed 42, identical across the two CLI runs). Custom Agent-TOC reaches \SDKCustomTOCAcc{} on Sonnet and \GPTFiveFourCustomTOCMatchedAcc{} on GPT-5.4 over the matched 50, within 4\,pp of the full-200 headline (\SonnetTOC) on Sonnet and within 1\,pp (\GPTFiveFourTOC) on GPT-5.4 (Table~\ref{tab:main-results}); the matched-50 numbers for the custom side were derived from the per-item failure-case package included in the artifact rather than from a fresh subsample run.

\section{Trajectory Length and Failure Mode}
\label{sec:appendix-trajectory}

Tool-augmented evaluation should include more than final accuracy because correctness may be purchased with long retries, repeated simulator executions, or high wall-clock cost. Table~\ref{tab:trajectory} reports trajectory statistics parsed from agent logs (wall-clock measured per question from the logged \texttt{run\_start} to \texttt{run\_end}). The aggregate medians (median 7--8 steps across the five completed agents) mask qualitatively different failure modes across capability tiers: top-tier agents (Opus, Sonnet) fail by step-budget exhaustion \emph{after} committing to an answer ($>$60\% of failures reach the 24-step ceiling); mid-tier GPT agents fail by premature commitment (failed-item median 7--9 steps, $<$10\% reach the ceiling); and the below-mid Gemini Flash 3 agent fails by step-budget exhaustion \emph{without} committing (117/200 reach the ceiling with no final-answer message). Step exhaustion looks the same in aggregate trajectory medians but means opposite things at the top and below-mid tiers; identical aggregate trajectory medians can therefore correspond to substantively different agent behaviors. The remainder of this appendix decomposes the trajectory-length distribution to surface this distinction further.

\begin{table}[h]
\centering
\small
\caption{Trajectory statistics for TOC-agent runs. Wall-clock columns are seconds per question. ``Runs'' counts PHREEQC executions per question; failed runs counts nonzero PHREEQC executions across the full run.}
\label{tab:trajectory}
\begin{tabular}{lcccccc}
\toprule
Model & Wall med. & Wall p90 & Steps med. & Runs med. & Runs mean & Failed runs \\
\midrule
Claude Opus 4.6 + TOC & \OpusWallMedianSeconds & \OpusWallPNinetySeconds & \OpusMedianSteps & \OpusMedianRuns & \OpusMeanRuns & \OpusFailedRuns \\
Claude Sonnet 4.6 + TOC & \SonnetWallMedianSeconds & \SonnetWallPNinetySeconds & \SonnetMedianSteps & \SonnetMedianRuns & \SonnetMeanRuns & \SonnetFailedRuns \\
GPT-5.4 + TOC & \GPTFiveFourWallMedianSeconds & \GPTFiveFourWallPNinetySeconds & \GPTFiveFourMedianSteps & \GPTFiveFourMedianRuns & \GPTFiveFourMeanRuns & \GPTFiveFourFailedRuns \\
GPT-5.2 + TOC & \GPTFiveTwoWallMedianSeconds & \GPTFiveTwoWallPNinetySeconds & \GPTFiveTwoMedianSteps & \GPTFiveTwoMedianRuns & \GPTFiveTwoMeanRuns & \GPTFiveTwoFailedRuns \\
GPT-5.1 + TOC & \GPTFiveOneWallMedianSeconds & \GPTFiveOneWallPNinetySeconds & \GPTFiveOneMedianSteps & \GPTFiveOneMedianRuns & \GPTFiveOneMeanRuns & \GPTFiveOneFailedRuns \\
Gemini 3 Flash Preview + TOC & \GeminiThreeFlashWallMedianSeconds & \GeminiThreeFlashWallPNinetySeconds & \GeminiThreeFlashMedianSteps & \GeminiThreeFlashMedianRuns & \GeminiThreeFlashMeanRuns & \GeminiThreeFlashFailedRuns \\
\bottomrule
\end{tabular}
\end{table}

\subsection{Step Count vs Accuracy on GPT-5.1 + TOC ($n$=600)}

Combining the three GPT-5.1 TOC reruns described in Appendix~\ref{sec:appendix-stability} (Table~\ref{tab:toc-rerun-variance}) gives 600 (item, rerun) observations with full per-item step counts and correctness labels. Table~\ref{tab:appendix-step-bucket} reports accuracy bucketed by step count.

\begin{table}[h]
\centering
\small
\setlength{\tabcolsep}{8pt}
\caption{GPT-5.1 + TOC: per-item accuracy bucketed by trajectory length, pooled across three independent reruns. Accuracy is approximately stable in the 5--11 step range (50--54\%), drops to 33--39\% in the 12--19 range, and collapses to 9\% in the 20--24 range. The six items reaching the 24-step ceiling are all incorrect.}
\label{tab:appendix-step-bucket}
\begin{tabular}{lccc}
\toprule
Step bucket & $n$ & Correct & Accuracy \\
\midrule
5--7 steps & 311 & 157 & 50.5\% \\
8--11 steps & 193 & 104 & 53.9\% \\
12--15 steps & 57 & 19 & 33.3\% \\
16--19 steps & 28 & 11 & 39.3\% \\
20--24 steps (incl. ceiling) & 11 & 1 & 9.1\% \\
\midrule
all & 600 & 292 & 48.7\% \\
\bottomrule
\end{tabular}
\end{table}

The bucket pattern indicates two regimes. Trajectories of 5--11 steps are productive search: the agent identifies the relevant TOC section, runs PHREEQC, inspects output, and commits to an answer at roughly the model's baseline accuracy. Trajectories of 12 or more steps correspond to retry behavior --- failed PHREEQC executions, repeated reads of the wrong section, or repeated tool calls without progress --- and accuracy collapses across this regime. The 24-step ceiling acts as a useful failure signal: for GPT-5.1 + TOC, every item that reaches the budget ceiling in our reruns is wrong.

\subsection{Two Failure Patterns Across Models}

For models other than GPT-5.1 we have per-item step counts only for items in the failure-case package (Section~\ref{sec:appendix-artifact}); successful-trajectory step counts are not available because the original \texttt{result.out}-bearing workspaces were not retained for correct items. This is sufficient to characterize \emph{how} each model's wrong items distribute across the step budget. Table~\ref{tab:appendix-failure-step-distribution} reports failed-item step statistics alongside the overall (correct + wrong) median from Table~\ref{tab:trajectory}.

\begin{table}[h]
\centering
\small
\setlength{\tabcolsep}{6pt}
\caption{Failed-item step distribution per TOC agent. ``Failed median/mean'' summarize the per-item step count on items the agent answered incorrectly. ``\% at 24-step ceiling'' is the share of failed items that reached the maximum step budget. The overall median (from Table~\ref{tab:trajectory}) reflects the full population including correct items.}
\label{tab:appendix-failure-step-distribution}
\begin{tabular}{lccccc}
\toprule
Model & Failed $n$ & Failed median & Failed mean & Overall median & \% at 24-step ceiling \\
\midrule
Opus + TOC & 31 & 24.0 & 19.0 & 8 & 67.7\% \\
Sonnet + TOC & 29 & 24.0 & 18.2 & 7 & 62.1\% \\
GPT-5.4 + TOC & 55 & 10.0 & 11.1 & 8 & 3.6\% \\
GPT-5.2 + TOC & 93 & 9.0 & 10.7 & 8 & 9.7\% \\
GPT-5.1 + TOC & 92 & 7.0 & 8.3 & 7 & 0.0\% \\
\bottomrule
\end{tabular}
\end{table}

The failure-step distribution divides cleanly into two regimes:

\paragraph{Retry-loop failures (top-tier).}
When Opus and Sonnet TOC agents fail, they typically exhaust the 24-step budget: failed-item median is 24, and 62--68\% of failures reach the step ceiling. The mean failed-item step count of 18--19 indicates that even the non-ceiling failures used a substantial fraction of the budget. These models keep working on a problem when the simulator output, the agent's interpretation of it, or the chosen answer choice is inconsistent --- they retry, re-read, and re-execute PHREEQC --- and the failure mode is therefore one of running out of attempts rather than committing to a wrong answer early.

\paragraph{Early-commit failures (mid-tier GPT).}
When GPT-5.1 and GPT-5.2 TOC agents fail, the failed-item step counts are very close to the overall medians (failed 7--9 vs overall 7--8), and almost no failures hit the step ceiling (0--10\%). These models do not enter long retry loops on wrong items. They reach a wrong answer in approximately the same number of steps that they would use to reach a right answer, then commit. The failure mode is misnavigation or misextraction followed by quick commitment, not budget exhaustion.

GPT-5.4 sits between these patterns: failed-item median 10 (slightly above its overall median of 8), with 3.6\% of failures reaching the ceiling. It exhibits more retry behavior than the mid-tier GPT pair but markedly less than the Anthropic agents.

\subsection{What This Adds to the Capability-Tier Reading}

The two failure patterns map onto the capability-tier framing of Section~\ref{sec:diagnostic} in a specific way: top-tier agents under TOC fail by \emph{persistence without success}, while mid-tier GPT agents fail by \emph{premature commitment}. Both fail; both contribute to the headline accuracy gap. But the implication for tool-augmented agent design differs:

\begin{enumerate}[leftmargin=*, itemsep=2pt]
    \item For top-tier agents, raising the step ceiling could in principle recover some retry-loop failures --- but at the cost of substantially higher wall-clock time and token cost on already-failing items.
    \item For mid-tier agents, raising the step ceiling would not help: their failures already commit before the ceiling. Recovering these failures requires interventions that change \emph{how} the agent navigates and commits (better TOC return formats, calibrated abstention, or richer tool feedback), not the budget for retries.
\end{enumerate}

This is a benchmark observation that aggregate trajectory medians cannot surface: identical median step counts across capability tiers conceal qualitatively different failure mechanisms.

\subsection{Automatic Log-Derived Failure Categorization}

Beyond the trajectory-length failure modes above, a coarser log-level inspection categorizes errors into four buckets that flag where to inspect trajectories before expert review: whether any PHREEQC execution failed during the trajectory, whether the trajectory had a successful PHREEQC run but produced a wrong final answer, whether the trajectory hit the step budget or otherwise produced no parseable final answer, and whether the final answer was simply unparseable.

\begin{table}[t]
\centering
\small
\setlength{\tabcolsep}{4pt}
\caption{Automatic upper-level failure categories from logs (all rows are Agent-TOC). Columns: ``PHREEQC failed'' = any PHREEQC execution failed during the trajectory; ``Run OK, wrong'' = successful run but wrong final answer; ``No final'' = step-budget hit or no parseable final-answer; ``Unparseable'' = final answer present but not parseable. These are not domain-root-cause labels: they identify where to inspect trajectories before expert review.}
\label{tab:auto-failure}
\begin{tabular}{lcccc}
\toprule
Model (+ TOC) & PHREEQC failed & Run OK, wrong & No final & Unparseable \\
\midrule
Opus 4.6 & \OpusErrSomeFailed & \OpusErrSuccessfulWrong & \OpusErrRetry & \OpusErrUnparseable \\
Sonnet 4.6 & \SonnetErrSomeFailed & \SonnetErrSuccessfulWrong & \SonnetErrRetry & \SonnetErrUnparseable \\
GPT-5.2 & \GPTFiveTwoErrSomeFailed & \GPTFiveTwoErrSuccessfulWrong & \GPTFiveTwoErrRetry & \GPTFiveTwoErrUnparseable \\
Gemini 3 Flash Preview & \GeminiThreeFlashErrSomeFailed & \GeminiThreeFlashErrSuccessfulWrong & \GeminiThreeFlashErrRetry & \GeminiThreeFlashErrUnparseable \\
\bottomrule
\end{tabular}
\end{table}

This automatic taxonomy already shows a useful distinction. For strong agents, many residual errors are not ordinary API failures or complete simulator failures. Instead, they often involve a successful PHREEQC execution followed by a wrong final answer, or a trajectory with at least one failed PHREEQC attempt followed by an incorrect recovery. Those cases require expert review (Section~\ref{sec:failure}) to separate plausible-but-wrong simulator inputs from output-navigation and answer-mapping errors. The below-mid Gemini Flash 3 row shows a qualitatively different signature: \GeminiThreeFlashErrRetry{} of its errors fall into the step/no-final category and another \GeminiThreeFlashErrUnparseable{} into no-parse, while only \GeminiThreeFlashErrSuccessfulWrong{} are successful-run-wrong --- i.e.\ the automatic taxonomy alone is sufficient to identify non-commit as the dominant residual failure mode, without needing expert review.

\section{Domain-Reviewed Failure Taxonomy}
\label{sec:appendix-failure-taxonomy}

\subsection{GPT-5.4 + Agent-TOC: Current-Evaluation Breakdown}

The domain reviewer's breakdown of all 55 GPT-5.4 + Agent-TOC errors on \BenchName{}:

\begin{table}[h]
\centering
\small
\caption{Domain-reviewed failure taxonomy for GPT-5.4~+~Agent-TOC on \BenchName{} (55/200 errors).}
\label{tab:appendix-gpt-failure}
\begin{tabular}{lcc}
\toprule
Failure category & Count & Share of errors \\
\midrule
High-confidence wrong build & 33 & 60\% \\
Wrong interpretation (right answer present, not used) & 12 & 22\% \\
Precision & 4 & 7\% \\
Bad simulation (PHREEQC failed) & 3 & 5\% \\
Low-confidence wrong build & 2 & 4\% \\
No question & 1 & 2\% \\
\midrule
Total errors & 55 & 100\% \\
\bottomrule
\end{tabular}
\end{table}

The reviewer also notes that ``the type of question gotten wrong is more widespread'' for GPT than for Opus, that ``the failure mode of right answer present but still answered wrong appears here where it doesn't in Opus,'' and that GPT ``has much more high-confidence wrong builds compared to Opus.''

\subsection{Categorical Descriptions (Paired Prior Domain Review)}

The categorical descriptions below are from a paired domain review of the same models on \BenchName{} under the prior PHREEQC-Agent evaluation; specific count breakdowns under that prior evaluation are in the released artifact (Appendix~\ref{sec:appendix-artifact}). Qualitative failure modes are stable across both evaluations.

\subsubsection{GPT 5.2 + Agent: Active Failure Modes}

Two dominant categories accounted for the majority of errors: (i) \textbf{wrong interpretation} (PHREEQC ran but the agent misread output or mapped to the wrong option), and (ii) \textbf{partial PHREEQC failure} (some attempts fail due to invalid inputs, sometimes followed by a successful run but still ending incorrect).

\paragraph{wrong\_interpretation.}
PHREEQC executes successfully and relevant values appear in \texttt{result.out}, but the agent reads the wrong field, the wrong timestep, or mis-maps a numeric result to the multiple-choice option. This is the dominant ``reasoning + extraction'' bottleneck and can be further divided into two categories: \textit{incorrect\_scripts}, where the agent writes a script that executes, but either makes a syntax error or incorrectly constructs the script in some manner that allows the output to run but does not accurately represent the scenario outlined in the problem statement, or \textit{incorrect\_interpretation}, where the script accurately reflects the scenario outlined in the question, but the agent interprets the output incorrectly. The vast majority of wrong interpretation errors for GPT 5.2 + Agent belong to the first category, primarily involving kinetics questions and questions involving gas phases, where the agent did not set up the problem correctly. In some scenarios, the agent would also assign the charge balancing adjustment to the pH parameter unprompted, causing the simulation to diverge from the problem statement. Another common failure point occurs when the agent writes a script that fails to correctly link different blocks together in multistep problems, thus creating disjointed outputs. \textit{Incorrect\_interpretation} events occurred for questions requiring counting the occurrence of a certain feature, such as ``How many species of Ca are present in a solution?''. The output file contains the correct results, but the model fails to return the correct information, revealing a critical limitation in the GPT 5.2 + Agent model.

\paragraph{some\_phreeqc\_failed and all\_phreeqc\_failed.}
The agent often generates invalid PHREEQC inputs (unsupported keywords, malformed blocks, unit mistakes), leading to parse/runtime errors. In \texttt{some\_phreeqc\_failed}, at least one execution succeeds but the overall trajectory still ends incorrect; in \texttt{all\_phreeqc\_failed}, no successful run occurs. The model has particularly poor performance in regards to kinetics questions. Kinetics questions involve a time dependent rate law that is accessed through the thermodynamic database. GPT 5.2 + Agent appears to not recognize the rate laws within the database, and thus struggles to write a functional script, either failing outright, running out of steps or timing out, or constructing its own nonsensical rate law, causing a \textit{wrong\_interpretation} failure.

\paragraph{max\_steps.}
The agent enters retry loops (iterative minor edits and re-runs) and exhausts the 24-step budget without producing a parseable final answer.

\paragraph{timeout/error.}
The per-question process exceeds the 300-second timeout or terminates unexpectedly.

\paragraph{no\_phreeqc\_run.}
A rare tool-bypass event: the agent answers without executing PHREEQC despite instructions.

\subsubsection{Claude Opus 4.6 + Agent: Failures in Detail}

Under the prior evaluation, Claude Opus 4.6 + Agent produced wrong-answer and no-answer trajectories with the following root-cause categories.

\paragraph{Bad PHREEQC input.}
The agent writes a PHREEQC input file that executes without errors but contains a subtle modeling flaw, producing plausible but incorrect output. Specific instances include omitting the \texttt{charge} keyword so PHREEQC accepts a user-specified pH instead of computing the equilibrium value, assigning charge balance to a lone anion, which PHREEQC silently ignores, and setting a kinetic rate constant too stiff for the integrator, causing the system to over-equilibrate in the first time step. In these cases, the simulation runs to completion and the agent trusts the incorrect output.

\paragraph{Wrong interpretation.}
PHREEQC output is correct, but the agent reasons to the wrong answer. Examples include misidentifying which output quantity corresponds to the question's target (e.g., confusing total alkalinity with HCO$_3^-$ molality) and misreading a positive mineral delta as ``zero dissolved.''

\paragraph{Precision limit.}
PHREEQC displayed a rounded version of the correct answer to the problem due to precision display limitations. The agent cannot distinguish the two answer choices from PHREEQC output alone. This question is unsolved by all ten methods.

\paragraph{PHREEQC retry loop.}
The agent repeatedly rewrites and re-executes PHREEQC input with varied configurations, exhausting all 24 steps without converging on a result that matches any answer choice.

\paragraph{Complex setup failure.}
The problem statement is challenging to convert to PHREEQC code due to complicated rate laws or sequential mixing-then-heating semantics, thus causing the system to write an erroneous script. Other questions with the same scenario were answered correctly, indicating that the system is capable of setting up those problem statements, but was unable to in this instance.

\section{Datasheet for \texorpdfstring{\BenchName}{PHREEQC-MCQ-200}}
\label{sec:appendix-datasheet}

This appendix documents the benchmark in the format of \citep{gebru2021datasheets}. We report only properties relevant to evaluation use; full reproducibility material is in the artifact (Section~\ref{sec:appendix-artifact}).

\subsection{Motivation}

\paragraph{Purpose.}
\BenchName{} was created to evaluate tool-augmented LLM agents on a deterministic scientific simulator under a unique-answer setting, with explicit support for retention, output-access ablation, scenario-clustered uncertainty, and failure-mode decomposition. The benchmark is designed as a measurement instrument for compound AI systems, not as a domain assessment of geochemistry.

\subsection{Composition}

\paragraph{Instances.}
Each instance is a four-option multiple-choice question with a unique simulator-derived target value, the four answer choices, and a designated correct option ($A$, $B$, $C$, or $D$). The benchmark contains \BenchN{} question items derived from \BenchScenarioN{} validated simulation scenario stems; multiple items may share a scenario stem when the simulation produces multiple distinct quantities (different output fields, time steps, or derived values). The truth-label distribution is A=63 (31.5\%), B=47 (23.5\%), C=48 (24.0\%), D=42 (21.0\%); the majority-label baseline is therefore 31.5\%.

\paragraph{Item content.}
Each item provides natural-language problem text describing a chemical scenario (e.g., dissolution of a mineral in a specified aqueous solution under specified thermodynamic conditions) and asks for a specific computed quantity (pH, saturation index, concentration of a particular species, time-resolved value, etc.). The agent receives the problem text and four answer choices; it does not receive a prewritten PHREEQC input file.

\paragraph{Format.}
JSONL with one item per line. Fields include \texttt{question}, four answer-choice fields, \texttt{answer} (the correct option), and metadata used by the evaluation harness.

\paragraph{Companion files.}
The benchmark is distributed with (i)~the author-supplied PHREEQC thermodynamic database used during item validation and required at evaluation time; (ii)~the agent system prompts (Direct, CoT, TOC, Raw) and tool contracts; (iii)~per-question chat logs and generated PHREEQC artifacts for failure cases; (iv)~scenario-stem groupings; and (v)~the analysis script that regenerates every table in this paper.

\paragraph{Output-size distribution.}
Simulator outputs are heavy-tailed and bimodal in our reference TOC run: \OutputLeHundredK{} of \texttt{result.out} files are at most 100k characters (95th percentile within this mode at $\sim$55k characters), and the remaining \OutputGtHundredK{} begin at $\sim$369k characters and extend to a maximum of \OutputMaxChars. The empty interval between modes is the rationale for the 100k raw-output cap (Section~\ref{sec:toc}). Output sizes per question are not a property of the dataset item; they are produced when an agent's PHREEQC input is executed, and across our five completed TOC runs the long-output count varies materially (23--41 items per run, union 45, intersection 15; Appendix~\ref{sec:appendix-cap-affected-per-model}). Bin-level analyses in Appendix~\ref{sec:appendix-output-bin-grid} use a single canonical reference run for bin assignment.

\subsection{Collection and Validation}

\paragraph{Authoring.}
Items were authored by domain experts in aqueous geochemistry and reviewed by additional independent reviewers.

\paragraph{Validation criteria.}
Each candidate item was retained only if (i)~the simulation setup executes successfully against the frozen author-supplied database, (ii)~the intended answer is uniquely determined from the PHREEQC output, (iii)~the four answer choices are numerically plausible (distractors are not trivially distinguishable from the correct option without simulation), and (iv)~the question statement is unambiguous to multiple independent reviewers.

\newpage

\end{document}

%% file: results_macros.tex
\newcommand{\BenchName}{PHREEQC-MCQ-200}
\newcommand{\BenchN}{200}
\newcommand{\BenchScenarioN}{21}
\newcommand{\BenchAlwaysA}{31.5\%}
\newcommand{\BenchAlwaysB}{23.5\%}
\newcommand{\BenchAlwaysC}{24.0\%}
\newcommand{\BenchAlwaysD}{21.0\%}
\newcommand{\BenchMajority}{31.5\%}
\newcommand{\BenchTruthCounts}{63/47/48/42}
\newcommand{\RawCapChars}{100{,}000}

\newcommand{\OutputLeHundredK}{81.5\%}
\newcommand{\OutputGtHundredK}{18.5\%}
\newcommand{\OutputPFiftyChars}{11.5k}
\newcommand{\OutputPNinetyChars}{480.1k}
\newcommand{\OutputMaxChars}{846.8k}
\newcommand{\RawCapUnaffectedN}{163}

\newcommand{\OpusOneShot}{42.0\%}
\newcommand{\OpusOneShotCI}{[35.4, 48.9]}
\newcommand{\SonnetOneShotCI}{[30.6, 43.9]}
\newcommand{\GPTFiveFourOneShotCI}{[21.3, 33.5]}
\newcommand{\GPTFiveTwoOneShotCI}{[21.8, 34.1]}
\newcommand{\GPTFiveOneOneShotCI}{[32.5, 45.9]}
\newcommand{\GeminiThreeFlashOneShotCI}{[32.0, 45.4]}
\newcommand{\OpusCoTCI}{[31.6, 44.9]}
\newcommand{\SonnetCoTCI}{[36.3, 49.9]}
\newcommand{\GPTFiveFourCoTCI}{[28.7, 41.8]}
\newcommand{\GPTFiveTwoCoTCI}{[28.3, 41.3]}
\newcommand{\GPTFiveOneCoTCI}{[28.7, 41.8]}
\newcommand{\GeminiThreeFlashCoTCI}{[33.9, 47.4]}
\newcommand{\OpusCoT}{38.0\%}
\newcommand{\SonnetOneShot}{37.0\%}
\newcommand{\SonnetCoT}{43.0\%}
\newcommand{\GPTFiveOneOneShot}{39.0\%}
\newcommand{\GPTFiveOneCoT}{35.0\%}
\newcommand{\GPTFiveTwoOneShot}{27.5\%}
\newcommand{\GPTFiveTwoCoT}{34.5\%}
\newcommand{\GPTFiveFourOneShot}{27.0\%}
\newcommand{\GPTFiveFourCoT}{35.0\%}
\newcommand{\GeminiFlashOneShot}{36.5\%}
\newcommand{\GeminiFlashCoT}{27.5\%}
\newcommand{\GeminiThreeFlashOneShot}{38.5\%}
\newcommand{\GeminiThreeFlashCoT}{40.5\%}

\newcommand{\OpusTOC}{83.5\%}      
\newcommand{\SonnetTOC}{84.0\%}        
\newcommand{\GPTFiveOneTOC}{54.0\%}    
\newcommand{\GPTFiveTwoTOC}{54.0\%}    
\newcommand{\GPTFiveFourTOC}{72.5\%}   

\newcommand{\GeminiThreeFlashTOC}{23.5\%}
\newcommand{\GeminiThreeFlashTOCMinusBestNoTool}{-17.0 pp}

\newcommand{\GeminiThreeFlashWallMedianSeconds}{113.0}
\newcommand{\GeminiThreeFlashWallPNinetySeconds}{158.6}
\newcommand{\GeminiThreeFlashMedianSteps}{24}
\newcommand{\GeminiThreeFlashMedianRuns}{6}
\newcommand{\GeminiThreeFlashMeanRuns}{5.3}
\newcommand{\GeminiThreeFlashFailedRuns}{130}
\newcommand{\GeminiThreeFlashScenarioMean}{23.4\%}
\newcommand{\GeminiThreeFlashScenarioCILow}{17.0\%}
\newcommand{\GeminiThreeFlashScenarioCIHigh}{29.9\%}
\newcommand{\GeminiThreeFlashScenarioAllCorrect}{0}
\newcommand{\GeminiThreeFlashScenarioAllWrong}{2}
\newcommand{\GeminiThreeFlashErrSomeFailed}{2}
\newcommand{\GeminiThreeFlashErrSuccessfulWrong}{1}
\newcommand{\GeminiThreeFlashErrRetry}{112}
\newcommand{\GeminiThreeFlashErrUnparseable}{38}

\newcommand{\GeminiTOCReadFileMedian}{9}

\newcommand{\GeminiTOCStepMedian}{24}

\newcommand{\GeminiRawReadFileMedian}{2}

\newcommand{\GeminiRawStepMedian}{18}

\newcommand{\SDKCustomTOCAcc}{80.0\%}

\newcommand{\SDKCustomRawAcc}{84.0\%}

\newcommand{\SDKSonnetAcc}{60.0\%}
\newcommand{\SDKSonnetTimeouts}{4}
\newcommand{\SDKSonnetMaxTurns}{6}
\newcommand{\SDKCustomTOCMinusSDK}{20.0\,pp}

\newcommand{\GPTFiveFourCustomTOCMatchedAcc}{72.0\%}

\newcommand{\GPTFiveFourCustomRawMatchedAcc}{70.0\%}

\newcommand{\GPTFiveFourCustomRawFiveHundredKMatchedAcc}{78.0\%}

\newcommand{\CodexGPTFiveFourAcc}{86.0\%}

\newcommand{\SonnetCustomTOCPerQK}{101}
\newcommand{\SonnetCustomRawPerQK}{258}
\newcommand{\SDKSonnetPerQK}{197}
\newcommand{\GPTFiveFourCustomTOCPerQK}{57}
\newcommand{\GPTFiveFourCustomRawPerQK}{72}
\newcommand{\GPTFiveFourCustomRawFiveHundredKPerQK}{129}
\newcommand{\CodexGPTFiveFourPerQK}{235}

\newcommand{\SonnetCustomTOCPerCorrectK}{126}
\newcommand{\SonnetCustomRawPerCorrectK}{307}
\newcommand{\SDKSonnetPerCorrectK}{328}
\newcommand{\GPTFiveFourCustomTOCPerCorrectK}{79}
\newcommand{\GPTFiveFourCustomRawPerCorrectK}{102}
\newcommand{\GPTFiveFourCustomRawFiveHundredKPerCorrectK}{165}
\newcommand{\CodexGPTFiveFourPerCorrectK}{273}

\newcommand{\OpusRaw}{83.0\%}        
\newcommand{\SonnetRaw}{83.5\%}
\newcommand{\GPTFiveOneRaw}{63.5\%}
\newcommand{\GPTFiveTwoRaw}{61.5\%}
\newcommand{\GPTFiveFourRaw}{70.0\%}

\newcommand{\GeminiThreeFlashRaw}{47.0\%}
\newcommand{\OpusTOCTokens}{23.76M}      
\newcommand{\SonnetTOCTokens}{22.14M}      

\newcommand{\GPTFiveOneTOCTokens}{9.51M}   

\newcommand{\GPTFiveTwoTOCTokens}{13.92M}  

\newcommand{\GPTFiveFourTOCTokens}{12.69M} 

\newcommand{\GPTFiveFourContextWindow}{1M}
\newcommand{\GPTFiveTwoContextWindow}{400k}
\newcommand{\GPTFiveOneContextWindow}{400k}

\newcommand{\GPTFiveOneRawFiveHundredKLargeAcc}{43.2\%}
\newcommand{\GPTFiveTwoRawFiveHundredKLargeAcc}{51.4\%}
\newcommand{\GPTFiveFourRawFiveHundredKLargeAcc}{59.5\%}
\newcommand{\GPTFiveOneRawLargeAcc}{35.1\%}
\newcommand{\GPTFiveTwoRawLargeAcc}{56.8\%}
\newcommand{\GPTFiveFourRawLargeAcc}{64.9\%}

\newcommand{\GPTFiveOneRawLargeTokensPerItem}{165.3k}
\newcommand{\GPTFiveTwoRawLargeTokensPerItem}{218.9k}
\newcommand{\GPTFiveFourRawLargeTokensPerItem}{155.3k}
\newcommand{\GPTFiveOneRawFiveHundredKLargeTokensPerItem}{290.2k}
\newcommand{\GPTFiveTwoRawFiveHundredKLargeTokensPerItem}{500.9k}
\newcommand{\GPTFiveFourRawFiveHundredKLargeTokensPerItem}{409.6k}

\newcommand{\OpusTOCFortyFiveAcc}{71.1\%}
\newcommand{\SonnetTOCFortyFiveAcc}{71.1\%}
\newcommand{\GPTFiveFourTOCFortyFiveAcc}{55.6\%}
\newcommand{\GPTFiveTwoTOCFortyFiveAcc}{48.9\%}
\newcommand{\GPTFiveOneTOCFortyFiveAcc}{40.0\%}
\newcommand{\OpusTOCFortyFiveTokensPerItem}{191K}
\newcommand{\SonnetTOCFortyFiveTokensPerItem}{193K}
\newcommand{\GPTFiveFourTOCFortyFiveTokensPerItem}{85K}
\newcommand{\GPTFiveTwoTOCFortyFiveTokensPerItem}{111K}
\newcommand{\GPTFiveOneTOCFortyFiveTokensPerItem}{101K}
\newcommand{\OpusRawFortyFiveAcc}{73.3\%}              
\newcommand{\SonnetRawFortyFiveAcc}{75.6\%}
\newcommand{\GPTFiveFourRawFortyFiveAcc}{60.0\%}
\newcommand{\GPTFiveTwoRawFortyFiveAcc}{57.8\%}
\newcommand{\GPTFiveOneRawFortyFiveAcc}{33.3\%}
\newcommand{\OpusRawFortyFiveTokensPerItem}{434K}      
\newcommand{\SonnetRawFortyFiveTokensPerItem}{546K}
\newcommand{\GPTFiveFourRawFortyFiveTokensPerItem}{147K}
\newcommand{\GPTFiveTwoRawFortyFiveTokensPerItem}{185K}
\newcommand{\GPTFiveOneRawFortyFiveTokensPerItem}{109K}

\newcommand{\OpusTOCMinusRawFortyFive}{-2.2 pp}        
\newcommand{\SonnetTOCMinusRawFortyFive}{-4.5 pp}      
\newcommand{\GPTFiveFourTOCMinusRawFortyFive}{-4.4 pp} 
\newcommand{\GPTFiveTwoTOCMinusRawFortyFive}{-8.9 pp}  
\newcommand{\GPTFiveOneTOCMinusRawFortyFive}{+6.7 pp}  

\newcommand{\LargeOutputN}{37}

\newcommand{\OpusKept}{70}        
\newcommand{\OpusGained}{97}      
\newcommand{\OpusLost}{14}        
\newcommand{\OpusRetention}{83.3\%}  
\newcommand{\SonnetKept}{64}        
\newcommand{\SonnetGained}{104}     
\newcommand{\SonnetLost}{10}        
\newcommand{\SonnetRetention}{86.5\%}  
\newcommand{\GPTFiveOneKept}{46}    
\newcommand{\GPTFiveOneGained}{62}  
\newcommand{\GPTFiveOneLost}{32}    
\newcommand{\GPTFiveOneRetention}{59.0\%}  
\newcommand{\GPTFiveTwoKept}{31}    
\newcommand{\GPTFiveTwoGained}{77}  
\newcommand{\GPTFiveTwoLost}{24}    
\newcommand{\GPTFiveTwoRetention}{56.4\%}  
\newcommand{\GPTFiveFourKept}{35}   
\newcommand{\GPTFiveFourGained}{110}  
\newcommand{\GPTFiveFourLost}{19}   
\newcommand{\GPTFiveFourRetention}{64.8\%}  

\newcommand{\OpusRawKept}{71}
\newcommand{\OpusRawGained}{95}
\newcommand{\OpusRawLost}{13}
\newcommand{\OpusRawRetention}{84.5\%}

\newcommand{\SonnetRawKept}{63}
\newcommand{\SonnetRawGained}{104}
\newcommand{\SonnetRawLost}{11}
\newcommand{\SonnetRawRetention}{85.1\%}

\newcommand{\GPTFiveOneRawKept}{52}
\newcommand{\GPTFiveOneRawGained}{75}
\newcommand{\GPTFiveOneRawLost}{26}
\newcommand{\GPTFiveOneRawRetention}{66.7\%}

\newcommand{\GPTFiveTwoRawKept}{34}
\newcommand{\GPTFiveTwoRawGained}{89}
\newcommand{\GPTFiveTwoRawLost}{21}
\newcommand{\GPTFiveTwoRawRetention}{61.8\%}

\newcommand{\GPTFiveFourRawKept}{34}
\newcommand{\GPTFiveFourRawGained}{106}
\newcommand{\GPTFiveFourRawLost}{20}
\newcommand{\GPTFiveFourRawRetention}{63.0\%}

\newcommand{\OpusTOCMinusRawRetention}{-1.2 pp}        
\newcommand{\SonnetTOCMinusRawRetention}{+1.4 pp}      
\newcommand{\GPTFiveFourTOCMinusRawRetention}{+1.8 pp} 
\newcommand{\GPTFiveTwoTOCMinusRawRetention}{-5.4 pp}  
\newcommand{\GPTFiveOneTOCMinusRawRetention}{-7.7 pp}  

\newcommand{\OpusWallMedianSeconds}{34.0}
\newcommand{\OpusWallPNinetySeconds}{144.0}
\newcommand{\OpusMedianSteps}{8}

\newcommand{\OpusMedianRuns}{1}
\newcommand{\OpusMeanRuns}{2.1}
\newcommand{\OpusFailedRuns}{61}
\newcommand{\SonnetWallMedianSeconds}{29.0}
\newcommand{\SonnetWallPNinetySeconds}{126.0}
\newcommand{\SonnetMedianSteps}{7}

\newcommand{\SonnetMedianRuns}{1}
\newcommand{\SonnetMeanRuns}{2.1}
\newcommand{\SonnetFailedRuns}{65}
\newcommand{\GPTFiveOneWallMedianSeconds}{15.0}
\newcommand{\GPTFiveOneWallPNinetySeconds}{31.2}
\newcommand{\GPTFiveOneMedianSteps}{7}

\newcommand{\GPTFiveOneMedianRuns}{1}
\newcommand{\GPTFiveOneMeanRuns}{1.8}
\newcommand{\GPTFiveOneFailedRuns}{124}
\newcommand{\GPTFiveTwoWallMedianSeconds}{20.0}
\newcommand{\GPTFiveTwoWallPNinetySeconds}{47.0}
\newcommand{\GPTFiveTwoMedianSteps}{8}

\newcommand{\GPTFiveTwoMedianRuns}{1}
\newcommand{\GPTFiveTwoMeanRuns}{2.0}
\newcommand{\GPTFiveTwoFailedRuns}{115}
\newcommand{\GPTFiveFourWallMedianSeconds}{19.0}
\newcommand{\GPTFiveFourWallPNinetySeconds}{36.1}
\newcommand{\GPTFiveFourMedianSteps}{8}

\newcommand{\GPTFiveFourMedianRuns}{1}
\newcommand{\GPTFiveFourMeanRuns}{1.7}
\newcommand{\GPTFiveFourFailedRuns}{74}

\newcommand{\OpusErrSomeFailed}{13}
\newcommand{\OpusErrSuccessfulWrong}{8}
\newcommand{\OpusErrRetry}{10}
\newcommand{\OpusErrUnparseable}{0}
\newcommand{\SonnetErrSomeFailed}{10}
\newcommand{\SonnetErrSuccessfulWrong}{9}
\newcommand{\SonnetErrRetry}{9}
\newcommand{\SonnetErrUnparseable}{1}
\newcommand{\GPTFiveTwoErrSomeFailed}{25}

\newcommand{\GPTFiveTwoErrSuccessfulWrong}{58}
\newcommand{\GPTFiveTwoErrRetry}{8}
\newcommand{\GPTFiveTwoErrUnparseable}{0}

\newcommand{\SonnetScenarioMean}{85.6\%}
\newcommand{\SonnetScenarioCILow}{77.8\%}
\newcommand{\SonnetScenarioCIHigh}{92.6\%}
\newcommand{\SonnetScenarioAllCorrect}{9}
\newcommand{\SonnetScenarioAllWrong}{0}
\newcommand{\OpusScenarioMean}{84.8\%}
\newcommand{\OpusScenarioCILow}{77.3\%}
\newcommand{\OpusScenarioCIHigh}{91.5\%}
\newcommand{\OpusScenarioAllCorrect}{10}
\newcommand{\OpusScenarioAllWrong}{0}
\newcommand{\GPTFiveFourScenarioMean}{73.6\%}
\newcommand{\GPTFiveFourScenarioCILow}{64.0\%}
\newcommand{\GPTFiveFourScenarioCIHigh}{82.8\%}
\newcommand{\GPTFiveFourScenarioAllCorrect}{5}
\newcommand{\GPTFiveFourScenarioAllWrong}{0}
\newcommand{\GPTFiveTwoScenarioMean}{51.0\%}
\newcommand{\GPTFiveTwoScenarioCILow}{41.7\%}
\newcommand{\GPTFiveTwoScenarioCIHigh}{60.3\%}
\newcommand{\GPTFiveTwoScenarioAllCorrect}{0}
\newcommand{\GPTFiveTwoScenarioAllWrong}{1}
\newcommand{\OpusDirectScenarioMean}{43.9\%}
\newcommand{\OpusDirectScenarioCILow}{36.6\%}
\newcommand{\OpusDirectScenarioCIHigh}{52.8\%}
\newcommand{\OpusDirectScenarioAllCorrect}{1}
\newcommand{\OpusDirectScenarioAllWrong}{0}

\newcommand{\OpusDirectPredDist}{60/19/55/66/0}
\newcommand{\OpusDirectLabelRange}{21.3--64.3\%}
\newcommand{\SonnetDirectPredDist}{24/22/101/53/0}
\newcommand{\SonnetDirectLabelRange}{12.8--68.8\%}
\newcommand{\GPTFiveFourDirectPredDist}{20/109/42/29/0}
\newcommand{\GPTFiveFourDirectLabelRange}{7.9--51.1\%}
\newcommand{\OpusTOCPredDist}{59/47/48/46/0}
\newcommand{\OpusTOCLabelRange}{83.3--85.7\%}
\newcommand{\SonnetTOCPredDist}{63/44/47/45/1}
\newcommand{\SonnetTOCLabelRange}{83.0--88.1\%}
\newcommand{\GPTFiveFourTOCPredDist}{61/56/48/34/1}
\newcommand{\GPTFiveOneDirectPredDist}{41/62/46/51/0}
\newcommand{\GPTFiveTwoDirectPredDist}{16/101/43/40/0}
\newcommand{\GPTFiveOneTOCPredDist}{63/52/31/37/17}
\newcommand{\GPTFiveTwoTOCPredDist}{60/41/31/53/15}
\newcommand{\GPTFiveOneDirectLabelRange}{31.2--52.4\%}
\newcommand{\GPTFiveTwoDirectLabelRange}{9.5--55.3\%}
\newcommand{\GPTFiveOneTOCLabelRange}{33.3--60.3\%}
\newcommand{\GPTFiveTwoTOCLabelRange}{39.6--65.1\%}
\newcommand{\GeminiThreeFlashDirectPredDist}{49/26/59/66/0}
\newcommand{\GeminiThreeFlashTOCPredDist}{22/9/11/8/150}
\newcommand{\GeminiThreeFlashDirectLabelRange}{25.5--47.6\%}
\newcommand{\GeminiThreeFlashTOCLabelRange}{16.7--33.3\%}
\newcommand{\GPTFiveFourTOCLabelRange}{66.7--85.1\%}

\newcommand{\TOCRerunNumSamples}{3}
\newcommand{\TOCRerunMaxSD}{4.1}     

\newcommand{\OpusTOCRerunValues}{86.0\%, 84.0\%, 84.0\%}
\newcommand{\OpusTOCRerunMean}{84.7\%}
\newcommand{\OpusTOCRerunSD}{1.2}
\newcommand{\OpusTOCRerunRange}{2.0}

\newcommand{\OpusTOCRerunAccDiff}{$+$2.5, $+$0.5, $+$0.5}
\newcommand{\OpusTOCRerunTokDiff}{$-$2.80, $+$0.70, $-$1.50}
\newcommand{\OpusTOCRerunTokSD}{1.77}
\newcommand{\OpusTOCRerunTokRange}{3.50}

\newcommand{\GPTFiveOneTOCRerunSD}{4.1}
\newcommand{\GPTFiveOneTOCRerunRange}{7.5}

\newcommand{\GPTFiveOneTOCRerunAccDiff}{$-$5.0, $+$1.5, $-$6.0}
\newcommand{\GPTFiveOneTOCRerunTokDiff}{$+$0.93, $+$1.08, $+$0.84}
\newcommand{\GPTFiveOneTOCRerunTokSD}{0.12}
\newcommand{\GPTFiveOneTOCRerunTokRange}{0.23}

\newcommand{\GPTFiveTwoTOCRerunSD}{0.8}
\newcommand{\GPTFiveTwoTOCRerunRange}{1.5}

\newcommand{\GPTFiveTwoTOCRerunAccDiff}{$+$2.0, $+$1.0, $+$0.5}
\newcommand{\GPTFiveTwoTOCRerunTokDiff}{$-$0.89, $-$0.82, $-$1.11}
\newcommand{\GPTFiveTwoTOCRerunTokSD}{0.15}
\newcommand{\GPTFiveTwoTOCRerunTokRange}{0.30}

\newcommand{\GPTFiveFourTOCRerunSD}{2.8}
\newcommand{\GPTFiveFourTOCRerunRange}{5.5}

\newcommand{\GPTFiveFourTOCRerunAccDiff}{$-$4.5, $-$1.0, $+$1.0}
\newcommand{\GPTFiveFourTOCRerunTokDiff}{$-$1.22, $-$0.90, $-$1.11}
\newcommand{\GPTFiveFourTOCRerunTokSD}{0.16}
\newcommand{\GPTFiveFourTOCRerunTokRange}{0.32}

\newcommand{\OpusTOCSmallOrig}{87.1\%, 15.17M}

\newcommand{\OpusTOCSmallRerunAccDiff}{$+$3.2, $+$0.6, $+$0.6}

\newcommand{\OpusTOCSmallRerunTokDiff}{$-$2.79, $+$0.70, $-$1.50}
\newcommand{\OpusTOCSmallRerunSD}{1.5}
\newcommand{\OpusTOCSmallRerunRange}{2.6}
\newcommand{\OpusTOCSmallRerunTokSD}{1.77}
\newcommand{\OpusTOCSmallRerunTokRange}{3.50}

\newcommand{\SonnetTOCSmallOrig}{87.7\%, 13.47M}

\newcommand{\SonnetTOCSmallRerunAccDiff}{$+$0.6, $+$0.6, $-$0.6}

\newcommand{\SonnetTOCSmallRerunTokDiff}{$+$0.16, $+$1.73, $+$0.45}
\newcommand{\SonnetTOCSmallRerunSD}{0.7}
\newcommand{\SonnetTOCSmallRerunRange}{1.3}
\newcommand{\SonnetTOCSmallRerunTokSD}{0.83}
\newcommand{\SonnetTOCSmallRerunTokRange}{1.56}

\newcommand{\GPTFiveFourTOCSmallOrig}{77.4\%, 8.87M}

\newcommand{\GPTFiveFourTOCSmallRerunAccDiff}{$-$5.8, $-$1.3, $+$1.3}

\newcommand{\GPTFiveFourTOCSmallRerunTokDiff}{$-$1.21, $-$0.89, $-$1.10}
\newcommand{\GPTFiveFourTOCSmallRerunSD}{3.6}
\newcommand{\GPTFiveFourTOCSmallRerunRange}{7.1}
\newcommand{\GPTFiveFourTOCSmallRerunTokSD}{0.16}
\newcommand{\GPTFiveFourTOCSmallRerunTokRange}{0.32}

\newcommand{\GPTFiveTwoTOCSmallOrig}{55.5\%, 8.93M}

\newcommand{\GPTFiveTwoTOCSmallRerunAccDiff}{$+$2.6, $+$1.3, $+$0.6}

\newcommand{\GPTFiveTwoTOCSmallRerunTokDiff}{$-$0.89, $-$0.82, $-$1.11}
\newcommand{\GPTFiveTwoTOCSmallRerunSD}{1.0}
\newcommand{\GPTFiveTwoTOCSmallRerunRange}{1.9}
\newcommand{\GPTFiveTwoTOCSmallRerunTokSD}{0.15}
\newcommand{\GPTFiveTwoTOCSmallRerunTokRange}{0.30}

\newcommand{\GPTFiveOneTOCSmallOrig}{58.1\%, 4.96M}

\newcommand{\GPTFiveOneTOCSmallRerunAccDiff}{$-$6.5, $+$1.9, $-$7.7}

\newcommand{\GPTFiveOneTOCSmallRerunTokDiff}{$+$0.92, $+$1.07, $+$0.84}
\newcommand{\GPTFiveOneTOCSmallRerunSD}{5.3}
\newcommand{\GPTFiveOneTOCSmallRerunRange}{9.7}
\newcommand{\GPTFiveOneTOCSmallRerunTokSD}{0.12}
\newcommand{\GPTFiveOneTOCSmallRerunTokRange}{0.23}

\newcommand{\SonnetTOCRerunValues}{84.5\%, 84.5\%, 83.5\%}        
\newcommand{\SonnetTOCRerunMean}{84.2\%}
\newcommand{\SonnetTOCRerunSD}{0.6}
\newcommand{\SonnetTOCRerunRange}{1.0}

\newcommand{\SonnetTOCRerunAccDiff}{$+$0.5, $+$0.5, $-$0.5}
\newcommand{\SonnetTOCRerunTokDiff}{$+$0.15, $+$1.72, $+$0.45}
\newcommand{\SonnetTOCRerunTokSD}{0.83}
\newcommand{\SonnetTOCRerunTokRange}{1.56}

\newcommand{\OpusTOCCapUnaffected}{87.7\%}      
\newcommand{\OpusRawCapUnaffected}{86.4\%}      
\newcommand{\OpusTOCMinusRawCapUnaffected}{$+$1.2\,pp}
\newcommand{\SonnetTOCCapUnaffected}{87.7\%}
\newcommand{\SonnetRawCapUnaffected}{85.9\%}
\newcommand{\SonnetTOCMinusRawCapUnaffected}{$+$1.8\,pp}
\newcommand{\GPTFiveFourTOCCapUnaffected}{76.1\%}
\newcommand{\GPTFiveFourRawCapUnaffected}{71.2\%}
\newcommand{\GPTFiveFourTOCMinusRawCapUnaffected}{$+$4.9\,pp}
\newcommand{\GPTFiveTwoTOCCapUnaffected}{55.2\%}
\newcommand{\GPTFiveTwoRawCapUnaffected}{62.6\%}
\newcommand{\GPTFiveTwoTOCMinusRawCapUnaffected}{$-$7.4\,pp}
\newcommand{\GPTFiveOneTOCCapUnaffected}{57.0\%}
\newcommand{\GPTFiveOneRawCapUnaffected}{69.9\%}
\newcommand{\GPTFiveOneTOCMinusRawCapUnaffected}{$-$12.9\,pp}

